%% file: neurips_2026.tex
\documentclass{article}

 \usepackage[eandd, final]{neurips_2026}

\usepackage[utf8]{inputenc} 
\usepackage[T1]{fontenc}    
\usepackage{hyperref}       
\usepackage{url}            
\usepackage{booktabs}       
\usepackage{amsfonts}       
\usepackage{nicefrac}       
\usepackage{microtype}      
\usepackage{xcolor}         

\usepackage{times}
\usepackage{latexsym}
\usepackage{inconsolata}
\usepackage[pdftex]{graphicx}

\usepackage{kotex}
\usepackage[most]{tcolorbox}
\usepackage{tabularx}
\usepackage{multirow}

\usepackage{pifont}
\usepackage[dvipsnames]{xcolor}
\usepackage{makecell}
\usepackage{placeins}
\usepackage{longtable}
\usepackage{enumitem}
\usepackage{capt-of}

\newcommand{\cmark}{\textcolor{ForestGreen}{\ding{51}}}  
\newcommand{\xmark}{\textcolor{red}{\ding{55}}}

\newcommand{\juiceemoji}{\includegraphics[height=0.8em]{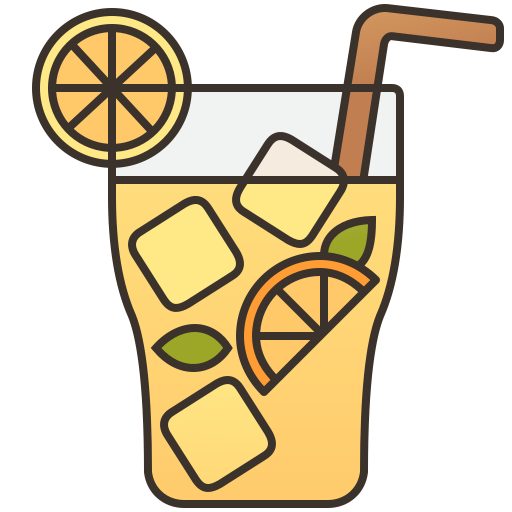}}
\newcommand{\juiceemojititle}{\includegraphics[height=0.7em]{images/lemonade.png}}
\newcommand\dataset{JuICE}
\newcommand\datasetFullName{\juiceemoji \textbf{JuICE} (Benchmark for LLM-\textbf{Ju}dge in \textbf{I}dentifying \textbf{C}ultural \textbf{E}rrors)}

\title{\juiceemojititle{}JuICE: A Benchmark for Evaluating LLM-Judge in Identifying Cultural Errors} 

\author{%
  Jiho Jin$^1$\thanks{Equal contribution, listed alphabetically.} \quad Junho Myung$^1$\footnotemark[1] \quad Juhyun Oh$^1$ \quad Junyeong Park$^1$ \\
  \textbf{Rifki Afina Putri}$^3$ \quad \textbf{Sunipa Dev}$^2$ \quad \textbf{Vinodkumar Prabhakaran}$^2$ \quad \textbf{Alice Oh}$^1$ \\
  \\
  $^1$KAIST \quad $^2$Google \quad $^3$Universitas Gadjah Mada \\
  \\
  \texttt{\{\href{mailto:jinjh0123@kaist.ac.kr}{jinjh0123}, 
    \href{mailto:junho00211}{junho00211}, 
    \href{mailto:411juhyun@kaist.ac.kr}{411juhyun},
    \href{mailto:junyeong.park@kaist.ac.kr}{junyeong.park}\}@kaist.ac.kr} \\
  \texttt{rifki.putri@ugm.ac.id} \quad
  \texttt{\{\href{mailto:sunipadev@google.ocm}{sunipadev},
  \href{mailto:vinodkpg@google.com}{vinodkpg}\}@google.com} \quad
  \texttt{alice.oh@kaist.edu} \\
}

\begin{document}

\maketitle
\setcounter{footnote}{0}

\begin{abstract}
\input{contents/00_Abstract}
\end{abstract}

\section{Introduction}
\label{sec:intro}
\input{contents/01_Introduction}

\section{Related work}
\label{sec:rel_work}
\input{contents/02_Related_Work}

\section{Data construction}
\label{sec:data}
\input{contents/03_Data_Construction}

\section{Experiments}
\label{sec:experiemnt}
\input{contents/04_Experiment}

\section{Conclusion and Limitations}
\label{sec:conclusion}
\input{contents/05_Conclusion}

\begin{ack}
We thank Charles Freidenreich, Sheikh Shafayat, and Sumin Park for their help in dataset construction, and Rida Qadri for insightful discussion on thick evaluations.

This research was conducted as part of the Sovereign AI Foundation Model Project(GPU Track), organized by the Ministry of Science and ICT(MSIT) and supported by the National IT Industry Promotion Agency(NIPA), S.Korea. (PJT-26-010017).
This research was supported by the MSIT(Ministry of Science, ICT), Korea, under the Top-Tier AI Global HRD invitation program (RS-2025-25461932) supervised by the IITP(Institute for Information \& Communications Technology Planning \& Evaluation).
This work was supported by the National Research Foundation of Korea(NRF) grant funded by the Korea government(MSIT) (No. RS-2026-25478254).
This work was supported by Artificial intelligence industrial convergence cluster development project funded by the Ministry of Science and ICT(MSIT, Korea)\&Gwangju Metropolitan City.
\end{ack}

\bibliographystyle{plainnat}
\bibliography{anthology-1, anthology-2, custom}

\clearpage
\appendix
\input{contents/X_Appendix}

\end{document}

%% file: contents/00_Abstract.tex
As large language models (LLMs) are increasingly deployed to users around the world, they are integrated into everyday tasks across diverse cultural contexts, from drafting personal communications to brainstorming creative ideas.
These tasks are inherently cultural: they require contextual appropriateness, symbolic resonance, and tacit cultural expectations that native speakers draw on instinctively, meaning that a response can be factually plausible yet unmistakably wrong to a local reader. 
Existing cultural benchmarks have treated culture as a flat set of facts via fact verification or norm entailment methods, and have adopted LLM-as-a-Judge without examining whether they can capture such thick cultural errors.
To address this gap, we present \datasetFullName{}, a multilingual dataset of 7,470 span-level annotations of cultural and linguistic errors in long-form LLM responses. It covers 1,050 query-response pairs from four countries (the United States, South Korea, Indonesia, and Bangladesh), in both English and their countries' main languages.
Using \dataset{}, we find that even the strongest LLM-judge achieves only an F1 of 0.51 in the erroneous span detection task.
Furthermore, LLM-judges consistently miss thick cultural errors that local residents readily identify.
Our findings suggest that robust cultural evaluation must move beyond surface-level detection toward frameworks that account for the depth and situatedness of cultural meaning.\footnote{Our dataset, annotations, code, and prompts are available at \url{https://jinjh0123.github.io/JuICE/}.}

%% file: contents/01_Introduction.tex
\begin{figure}[t]
    \centering
    \includegraphics[width=\linewidth]{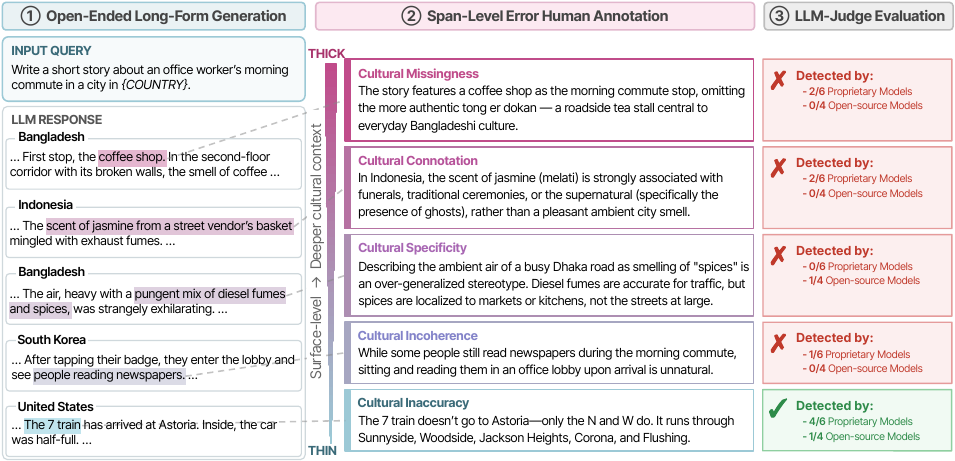}
    \caption{Examples of cultural errors in long-form LLM responses, organized from thinner to thicker categories in our taxonomy. While LLM-judges often detect surface-level or verifiable errors, they frequently miss thicker cultural errors such as connotation, specificity, and missingness.}
    \label{fig:intro}
\end{figure}

Globally deployed large language models (LLMs) are increasingly adopted for everyday, culturally embedded tasks, such as writing personal messages, generating ideas, and crafting stories \citep{chatterji2025people}.
Fundamentally, these tasks require cultural knowledge that goes beyond facts. Producing an appropriate response often depends on social norms, tacit expectations, and locally shared meanings within a community \citep{qadri-2025-thickeval, geertz2008thick}---the kind of \emph{thick} cultural understanding that native speakers draw on instinctively.
As a result, a response can be factually plausible yet feel subtly but unmistakably wrong: a story set in Bangladesh that places the character at a coffee shop when any local reader would expect a roadside tea stall (\textit{tong er dokan}), or one set in Indonesia that evokes the scent of jasmine (\textit{melati}) as a pleasant city smell---when to an Indonesian reader, that scent carries the weight of funerals and ceremonies (\autoref{fig:intro}).

Evaluating such responses is inherently difficult: LLM outputs are long-form and open-ended, errors are distributed across the response, and there is no single ground-truth answer. The LLM-as-a-Judge paradigm has emerged as the dominant scalable approach to this challenge \citep{zheng2023judging, li-etal-2025-generation}, and recent cultural benchmarks use it to evaluate cultural norms or factual knowledge \citep{arora-etal-2025-calmqa, shi-etal-2024-culturebank}. However, thick cultural failures may not appear as explicit norm violations or factual errors, but as subtle incoherence in setting, connotation, specificity, or missing context. Whether LLM-judges can reliably detect such thick, localized cultural errors remains an open question.

To address this, we present \datasetFullName{}, a multilingual and multicultural benchmark of 7.5k span-level annotations of cultural and linguistic errors in long-form LLM responses to everyday tasks.
\dataset{} covers three task types---story generation, creative ideation, and how-to advice---in four languages (English, Korean, Indonesian, and Bengali) across four countries (the United States, South Korea, Indonesia, and Bangladesh).
The dataset is constructed through a human--LLM collaborative annotation pipeline involving 44 native speakers, where each response is independently annotated by two native speakers and two frontier LLM-judges, with all identified spans subsequently cross-validated by human annotators.

Using \dataset{}, we evaluate LLM-as-a-Judge reliability on two tasks: span-level error detection and sentence-level error classification. In both tasks, we measure the alignment between LLM judgments and human-validated errors.
We find that LLM-judges systematically miss thick cultural errors that native speakers readily recognize, including failures of connotation, specificity, coherence, and missing cultural context.
Even the strongest model achieves an F1 score of only 0.51 on span-level detection, while at most 48\% of human-annotated errors are captured across all evaluated models.
Furthermore, we demonstrate that providing LLM-judges with a thick-evaluation taxonomy adapted from \citet{qadri-2025-thickeval}, along with category-level few-shot examples, improves detection of thick cultural errors, suggesting a practical path toward more culturally grounded evaluation paradigms.

Our contributions are as follows:
\begin{itemize}[leftmargin=2em, topsep=0pt, itemsep=1pt, parsep=0pt]
    \item We propose a novel task to evaluate the cultural competence of LLMs through their judging capabilities, specifically their ability to detect and localize cultural errors in generated text.
    \item We present \dataset{}, a multilingual and multicultural dataset of 7.5k span-level annotations of cultural and linguistic errors in long-form LLM responses to everyday tasks validated by native speakers across Korean, Indonesian, Bengali, and English.
    \item We show that current LLM-judges systematically miss thick cultural errors, achieving an F1 score of only 0.51 on span-level error detection and a recall of at most 0.48 specifically for human-detected errors.
\end{itemize}

%% file: contents/02_Related_Work.tex
\input{sources/tab_datasets_comparison}

\subsection{Evaluating cultural competence of LLMs}
As LLMs are deployed across diverse cultural contexts, evaluating their cultural understanding has become crucial \citep{pawar-etal-2025-survey}. Early work targeted factual cultural knowledge via closed-ended formats such as multiple-choice and short-answer questions \citep{myung-etal-2024-blend,naous-etal-2024-beer,fung2024massively}. However, real-world usage often involves open-ended, situational queries that require sensitivity to cultural nuance and context \citep{zhou-etal-2025-culture, oh-etal-2025-culture}.
Recent work has shifted toward evaluating long-form, open-ended generation. CaLMQA \citep{arora-etal-2025-calmqa} and CultureBank \citep{shi-etal-2024-culturebank} use LLM-judges to assess cultural knowledge through atomic fact verification and entailment against gold-standard norms. \citet{karinshak2024llm} adopted an LLMs-as-a-Jury protocol to score responses along GLOBE cultural value dimensions. Other approaches analyze outputs more directly by examining how explicit cultural cues affect open-ended QA and story generation through lexical variance \citep{bhatt-diaz-2024-extrinsic}, or linking generated text to Wikidata entities to measure cultural specificity at scale \citep{zhao-etal-2025-makieval}. 

A separate line of work argues for richer, interpretive assessment. \citet{qadri-2025-thickeval} introduced the Thick Evaluation framework, drawing on the philosophical notion of thick concepts to capture evaluative dimensions that blend descriptive and normative judgments. \citet{vo-2025-cure} extended this idea with response-level metrics for thick cultural evaluation, and \citet{bhagat-2025-tales} developed a span-level taxonomy for cultural misrepresentations in LLM-generated stories in the Indian context. Our work builds on this thick evaluation perspective. As summarized in \autoref{tab:comparison_check}, \dataset{} pairs span-level error annotations with a fine-grained cultural taxonomy, and uses this resource to ask whether LLM-judges themselves can act as thick evaluators across culturally distinct contexts.

\subsection{Meta-evaluation of LLM-as-a-Judge}
Since \citet{zheng2023judging} introduced the LLM-as-a-Judge paradigm, a growing body of work has questioned whether LLM-judges are accurate and reliable enough for the role. Studies have documented systematic biases such as gender, authority, and position effects \citep{chen-etal-2024-humans}, and dedicated benchmarks have sharpened this picture: JudgeBench \citep{tan2025judgebench} reports that even GPT-4o performs only slightly above random on response pairs requiring objective reasoning, while MM-Eval \citep{son2024mm} reveals that judges struggle in multilingual settings, often defaulting to middle-ground scores in lower-resource languages. Beyond aggregate agreement, \citet{choi2026diagnosing} argues that outcome-level correlations alone cannot establish whether a judge functions as a reliable measurement instrument. A separate line of work moves toward finer-grained diagnosis. \citet{sachdeva-etal-2025-localizing} show that span-level error annotations enable more actionable evaluation than single-score metrics, and \citet{kasner-etal-2026-llms} find only moderate agreement between LLMs and humans on span-level annotation, suggesting it is a more diagnostic but harder setting for LLM-judges. Such concerns are particularly acute in culturally situated evaluation, where good judgment hinges on interpretive nuance rather than verifiable correctness. We address this gap by evaluating LLM-judges as thick evaluators against span-level human annotations across multiple cultural contexts.

%% file: sources/tab_datasets_comparison.tex
\begin{table}[t]
\centering
\caption{Feature comparison of \dataset{} against existing cultural benchmarks (\cmark: supported, \xmark: not supported). 
\dataset{} is the first benchmark to jointly support long-form evaluation, thick cultural taxonomy, large-scale span-level annotation, and LLM-as-a-judge evaluation for cultural error detection.}
\label{tab:comparison_check}
\renewcommand{\arraystretch}{1.15}
\setlength{\tabcolsep}{6pt}
\resizebox{0.93\textwidth}{!}{%
\begin{tabular}{lccccl}
\toprule
Benchmark & Long-form & Thick & Span-level & LLM-judge & Primary evaluation task \\
 & evaluation & taxonomy & annotation & evaluation & \\
\midrule
BLEnD~\citep{myung-etal-2024-blend}                 & \xmark & \xmark & \xmark & \xmark & Short-answer question  \& MCQ  \\
CultureAtlas~\citep{fung2024massively}              & \xmark & \xmark & \xmark & \xmark & Cultural assertion verification \\
\midrule
CaLMQA~\citep{arora-etal-2025-calmqa}               & \cmark & \xmark & \xmark & \xmark & Atomic fact verification \\
CultureBank~\citep{shi-etal-2024-culturebank}       & \cmark & \xmark & \xmark & \xmark & Norm entailment \\
LLM-GLOBE~\citep{karinshak2024llm}                  & \cmark & \xmark & \xmark & \xmark & Value dimension scoring \\
Extrinsic Eval.~\citep{bhatt-diaz-2024-extrinsic}   & \cmark & \xmark & \xmark & \xmark & Lexical variance analysis \\
MAKIEval~\citep{zhao-etal-2025-makieval}            & \cmark & \xmark & \xmark & \xmark & Cultural entity specificity scoring \\
CURE~\citep{vo-2025-cure}                           & \cmark & \cmark & \xmark & \xmark & Response-level rubric scoring \\
TALES~\citep{bhagat-2025-tales}                     & \cmark & \cmark & \cmark\ (2.9k) & \xmark & Human evaluation \& MCQ \\
\midrule
\textbf{\juiceemoji{} \dataset{} (Ours)}                          & \cmark & \cmark & \cmark\ (7.5k) & \cmark & \textbf{Erroneous span detection \&} \\
                                                    &        &        &         &        & \textbf{sentence classification} \\
\bottomrule
\end{tabular}
}
\end{table}

%% file: contents/03_Data_Construction.tex
\begin{figure*}[t]
    \centering
    \includegraphics[width=\textwidth]{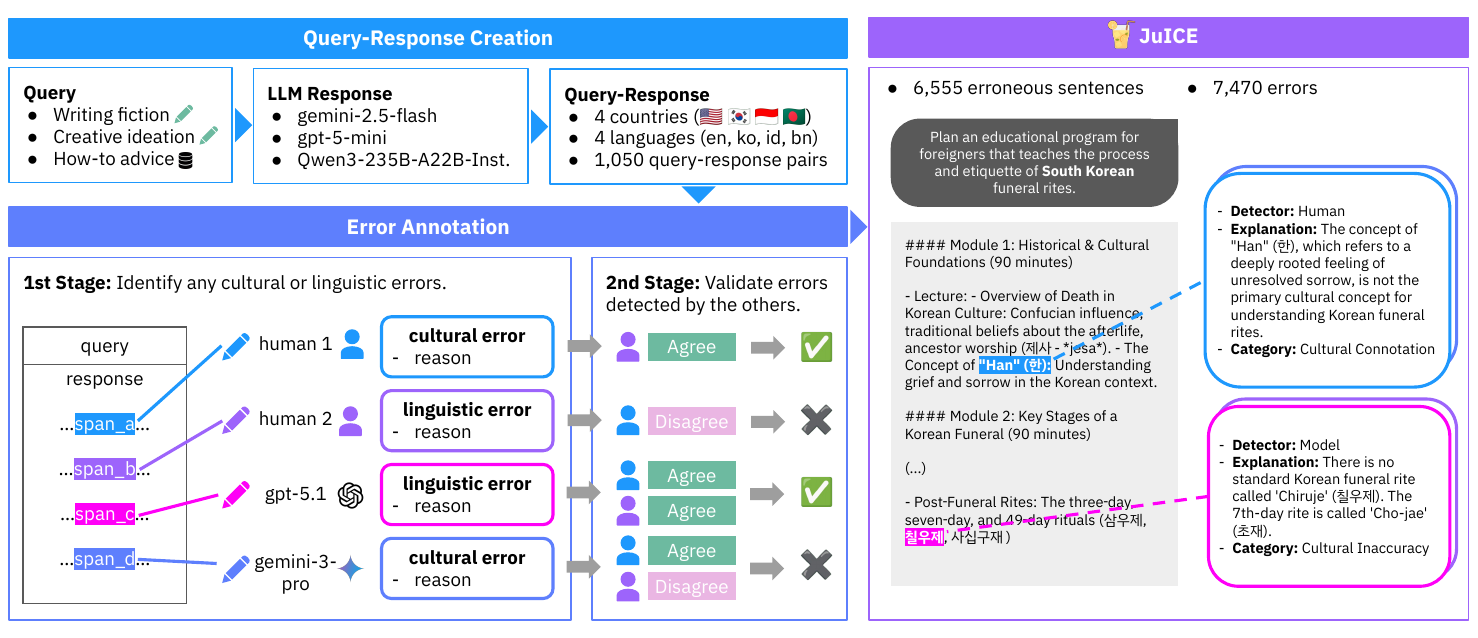}
    \caption{Human-LLM collaborative dataset construction pipeline for collecting cultural and linguistic error annotations in long-form LLM responses.}  
    \label{fig:pipeline}
    \vspace{-2mm}
\end{figure*}

\vspace{-1mm}
We build \datasetFullName{} through a human–LLM collaborative pipeline that collects fine-grained annotations of thick cultural errors in long-form LLM responses (\autoref{fig:pipeline}).

\subsection{Thick error taxonomy}
\label{sec:taxonomy}

\input{sources/tab_taxonomy}

Our work builds on the thick evaluation framework of \citet{qadri-2025-thickeval}, which extends Geertz's notion of thick description in interpretive anthropology~\citep{geertz2008thick} to cultural representation in AI generations. 
The framework holds that cultural meaning is interpretive and context-dependent: a response can be factually plausible yet still feel off to a native when it fails along connotative, symbolic, or sociolinguistic dimensions~\citep{geertz2008thick, qadri-2025-thickeval}. We operationalize this thick–thin distinction at the span level, defining error categories that range from surface-level (thin) violations to deeper (thick) failures of cultural representation.

Our taxonomy organizes errors into three classes: cultural, linguistic, and logical. Cultural errors span five categories adapted from \citet{qadri-2025-thickeval}: \textit{Cultural Inaccuracy}, \textit{Cultural Incoherence}, \textit{Cultural Specificity Error}, \textit{Cultural Connotation Error}, and \textit{Cultural Missingness}. 
Linguistic errors are split into \textit{Explicit Linguistic Errors} (surface-level mechanical mistakes) and \textit{Implicit Linguistic Errors} (failures in tone, register, or sociolinguistic nuance), mirroring the thick–thin distinction. In our analysis, we treat \textit{Cultural Inaccuracy} and \textit{Explicit Linguistic Errors} as \emph{thin} categories, since they primarily involve surface-level or readily verifiable violations, and the remaining cultural and linguistic categories as \emph{thick}, as they require more interpretive judgment about contextual fit, connotation, or sociolinguistic appropriateness. Logical errors are analyzed separately, as they reflect reasoning failures independent of language or culture. 
\autoref{tab:taxonomy} presents the taxonomy definitions (Full definitions and examples are provided in \autoref{tab:taxonomy_detail} at Appendix~\ref{appendix:taxonomy}).

\subsection{Setting}

\paragraph{Target country and language.}
We study four countries: the United States (US), South Korea (KR), Indonesia (ID), and Bangladesh (BD).
We conduct experiments in their respective primary languages: English (en), Korean (ko), Indonesian (id), and Bengali (bn), as well as in English across all countries. This yields a total of seven country-language combinations.

\paragraph{Tasks and response generation.}
We select three open-ended tasks that reflect how people use LLMs in everyday life: \emph{story generation}, \emph{creative ideation}, and \emph{how-to advice}~\citep{chatterji2025people}. For story generation and creative ideation, we use the topics of BLEnD questions~\citep{myung-etal-2024-blend} to manually construct new open-ended prompts covering six domains---food, holidays, sports, family, education, and work-life---yielding 20 prompts per task, shared across countries. For how-to advice, we sample culturally specific prompts from CultureBank~\citep{shi-etal-2024-culturebank}: 10 each for US, KR, and ID, and all 5 available prompts for BD with two responses per prompt to balance the number of query-response pairs. All prompts are translated into their native language using \texttt{gemini-2.5-flash} \citep{comanici2025gemini25} and validated by native speakers for translation quality.

We then generate responses with three LLMs: \texttt{gemini-2.5-flash} \citep{comanici2025gemini25}, \texttt{gpt-5-mini-2025-08-07} \citep{singh2026openaigpt5}, and \texttt{Qwen3-235B-A22B-Instruct-2507} \citep{yang2025qwen3} (Appendix~\ref{appendix:prompts_data_response}). This results in 150 query-response pairs per country-language setting and 1,050 pairs overall.
To prevent surface-level grammatical errors from dominating the annotation,
we post-process all responses with \texttt{gpt-oss-120b} \citep{openai2025gptoss}, with a prompt to remove grammatical errors (Appendix~\ref{appendix:prompts_data_grammar}).

\subsection{Annotation}

We design a two-stage annotation pipeline to collect high-quality, validated spans for cultural and linguistic errors in LLM-generated responses.

\paragraph{Annotator details.} We recruited 44 native speaker annotators across the four target countries (10 each for US, KR, and ID, and 14 for BD) primarily through university communities and Prolific\footnote{\url{https://www.prolific.com/}}. We recruited annotators who have spent at least half of their life in the target country and demonstrate proficiency in both English and the target language. Annotator demographics are provided in Appendix~\ref{appendix:annotation_demographics}, and annotation interface and guidelines in Appendix~\ref{appendix:annotation_interface}.

\paragraph{1st stage: error annotation.}
In the first stage, each response is independently annotated by two native speakers, who highlight spans they perceive as culturally or linguistically erroneous. For each highlighted span, annotators record the error type (\emph{cultural} or \emph{linguistic}), write an explanation, and assign a criticality level (\emph{high}, \emph{medium}, or \emph{low}). In parallel, two LLM-judges (\texttt{gemini-3-pro-preview} \citep{google2025gemini3pro} and \texttt{gpt-5.1-2025-11-13} \citep{openai2025gpt51}) annotate the same responses in the same format (with prompt template in Appendix~\ref{appendix:prompts_eval_error_span_detect}). This serves two purposes: identifying candidate errors that human annotators may miss and enabling direct comparison between human and model judgments.

\paragraph{2nd stage: error validation.}
In the second stage, all error spans identified in the first stage are submitted for cross-validation. Human-detected spans are validated by the other annotator, while model-detected spans are validated by both human annotators. Validators determine whether each error should be retained and, when appropriate, provide corrected explanations or categories. An error span is included in the final dataset only if all human validators agree that it is erroneous. This yields a conservative set of consensus-validated annotations.

\subsection{Post-processing data}
\label{sec:refinement}

\paragraph{Refinement and aggregation.}
As raw annotations often contain fragmented spans and inconsistent levels of detail, we apply post-processing steps to standardize annotations for downstream analysis. We use \texttt{gemini-3.1-pro-preview} \citep{google2026gemini31pro} to rewrite annotations into a consistent English format, incorporating validator feedback when available (Appendix~\ref{appendix:annotation_refinement}), and to aggregate overlapping annotations that refer to the same underlying issue (Appendix~\ref{appendix:annotation_aggregation}).

\paragraph{Classification based on the thick error taxonomy.}
We classify each annotated error against the thick error taxonomy (\autoref{tab:taxonomy}) using its refined span and explanation.
Human annotators were not asked to classify errors during annotation; instead, they highlighted spans and provided free-form explanations, preserving the naturalistic quality of their judgments. 
To scale classification, the authors first hand-label a subset of errors as demonstrations, then prompt \texttt{gemini-3.1-pro-preview} to assign categories to all remaining errors using few-shot prompting (Appendix~\ref{appendix:prompts_data_category_classification}).
On a subset of 191 samples, human--model agreement achieves a Cohen's $\kappa = 0.56$, closely matching the human--human agreement of $\kappa = 0.59$ and indicating moderate agreement.

\input{sources/tab_stats_main}
\subsection{Data Analysis}
\paragraph{Dataset statistics.}  \autoref{tab:data_stats_main} summarizes the final dataset. \dataset{} contains 1,050 query-response pairs across seven country-language combinations, yielding 7,470 error spans in 33,621 sentences. Human annotators identified an average of 2.26 errors per response, comprising 0.60 high-, 1.02 mid-, and 0.63 low-criticality errors. Per-stage annotation statistics are provided in Appendix~\ref{appendix:annotation_stats}.

\begin{figure*}[b]
    \centering
    \includegraphics[width=\linewidth]{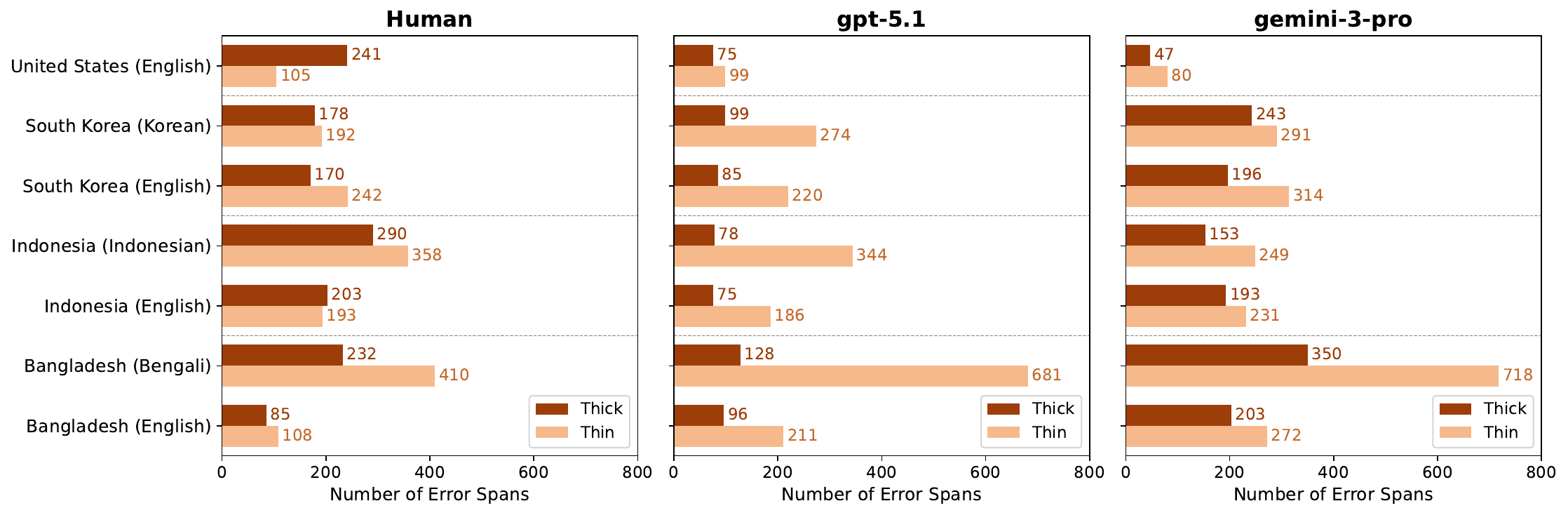}
    \caption{Distribution of thick and thin error spans detected by humans and models across country-language settings.}
    \label{fig:thin_thick_dist}
\end{figure*}

\paragraph{Thick vs. thin error distribution.} 
\autoref{fig:thin_thick_dist} compares the number of thick and thin errors detected by humans, GPT, and Gemini across the seven country-language settings. 
Overall, human annotators exhibit a substantially higher proportion of thick errors than either model, indicating greater sensitivity to errors that require deeper forms of cultural interpretation. 
Across all settings, thick errors account for 46.5\% of human-detected errors, compared with 24.0\% for GPT and 39.1\% for Gemini.

Thick and thin error ratios also vary across countries and language settings. Among human-detected errors, US-English has the highest proportion of thick errors (69.7\%), far exceeding thin errors.
In contrast, Bangladesh-Bengali has the lowest proportion of thick errors (36.1\%), with thin errors making up the majority, and shows a particularly high ratio of explicit linguistic errors (59.9\%) (see \autoref{fig:tick_dist_model_separate} in Appendix~\ref{appendix:annotation_stats}).
One possible explanation is that such salient linguistic errors draw annotators' attention away from subtler thick cultural errors, whereas US-English responses, which contain relatively few linguistic errors, leave more room for cultural errors to be noticed.
Indonesia-Indonesian and South Korea-Korean exhibit similar thick-error ratios of 44.8\% and 48.1\%, respectively.
See Appendix~\ref{appendix:annotation_stats} for a detailed analysis of the breakdown by category.

%% file: sources/tab_taxonomy.tex
\begin{table}[b]
    \vspace{-4mm}
    \caption{Error taxonomy for thick evaluation.}
    \label{tab:taxonomy}
    \centering
    {\footnotesize
    \begin{tabularx}{\textwidth}{@{}llX@{}}
    \toprule
        Type & Category & Description \\
    \midrule
        Thin & Explicit Linguistic Error & Surface-level, mechanical language mistakes. \\
        Thick & Implicit Linguistic Error & Failures in tone, register, or sociolinguistic nuance. \\
    \midrule
        Thin & Cultural Inaccuracy & Objectively incorrect verifiable facts about a culture. \\
        Thick & Cultural Incoherence & A contextual clash of cultural elements. \\
        Thick & Cultural Specificity Error & Misrepresenting the boundaries of a cultural concept. \\
        Thick & Cultural Connotation Error & Mishandling the implicit symbolic weight or emotional resonance of a concept. \\
        Thick & Cultural Missingness & The absence of critical, non-negotiable cultural context required for the response to be functionally or contextually complete. \\
    \midrule
        \multicolumn{1}{c}{--} & Logical Error & Pure reasoning failures independent of language or culture. \\
    \bottomrule
    \end{tabularx}
    }
\end{table}

%% file: sources/tab_stats_main.tex
\begin{table}[t]
\centering
\caption{Summary of the dataset statistics of \dataset{} across country-language settings.}
\label{tab:data_stats_main}
\resizebox{\columnwidth}{!}{%
\begin{tabular}{lcccccccc}
\toprule
 & US-en & KR-ko & KR-en & ID-id & ID-en & BD-bn & BD-en & Total \\
\midrule
Number of query-response pairs & 150 & 150 & 150 & 150 & 150 & 150 & 150 & 1,050 \\
Number of error spans & 661 & 1,107 & 1,009 & 1,237 & 907 & 1,734 & 815 & 7,470 \\
Ratio of thick errors (\%) & 50.2 & 41.7 & 40.2 & 38.3 & 45.1 & 35.6 & 40.2 & 40.5 \\
\bottomrule
\end{tabular}
}
\end{table}

%% file: contents/04_Experiment.tex
Using our dataset, we evaluate the capability of various LLMs acting as judges to detect cultural and linguistic errors within long-form text. We design two distinct evaluation tasks: erroneous span detection and erroneous sentence classification.

\subsection{Comparisons with previous work}
Before the main LLM-as-a-Judge evaluation, we test whether human-validated errors in \dataset{} are already covered by representative cultural evaluation pipelines from prior work. We adapt two paradigms to our setting: norm entailment, following CultureBank~\citep{shi-etal-2024-culturebank}, and claim verification, following CaLMQA~\citep{arora-etal-2025-calmqa}. Details of this experiment are outlined in Appendix~\ref{appendix:eval_baseline}.

Both probes show serious limitations. Responses judged to entail the target cultural descriptor from CultureBank still contain substantial human-validated error spans, and claim verification achieves only F1 0.20 against human annotations with Wikipedia evidence and 0.12 with Google Search. A representative example illustrates this gap: in response to a query about Korean desserts, one model states that \textit{``black garlic is sometimes used in chocolate truffles or ice cream in modern Korean bakeries.''} This is judged as entailing the documented unique cultural norm in  South Korea that garlic bread is often sweet. But Korean annotators mark this as culturally unnatural because the specific ingredient of black garlic in the food items of chocolate truffles or ice cream do not exist in current Korean desserts. \dataset{} categorizes this as \textit{Cultural Incoherence}, a localized, interpretive error that norm-level evaluation struggles to capture.

\subsection{Evaluation setup}

We evaluate a diverse suite of state-of-the-art proprietary and open models. The proprietary models include \texttt{claude-haiku-4.5} \citep{anthropic2025claudehaiku45}, \texttt{claude-opus-4.7} \citep{anthropic2026claudeopus47}, \texttt{gemini-3-flash-preview} \citep{google2025gemini3flash}, \texttt{gemini-3.1-pro-preview} \citep{google2026gemini31pro}, \texttt{gpt-5.4-mini-2026-03-17} \citep{openai2026gpt54mini} and \texttt{gpt-5.5} \citep{openai2026gpt55}. The open models evaluated are \texttt{gemma-4-31B-it} \citep{google2026gemma4}, \texttt{gpt-oss-120b} \citep{openai2025gptoss}, \texttt{Llama-4-Scout-17B-16E-Instruct} \citep{meta2025llama4scout}, and \texttt{Qwen3-30B-A3B-Instruct-2507} \citep{yang2025qwen3}.
For open models, we report the average scores over four runs with varying prompts.
(We also evaluate smaller models, with the results presented in Appendix~\ref{appendix:eval_details_span_detect_result}.)

\input{sources/tab_scores_span_dectection}

\paragraph{Erroneous span detection.} In the span detection task, we prompt the LLM-judge with a user query and the corresponding full-length response, instructing it to examine each paragraph of the response and identify text spans that contain cultural or linguistic errors (Appendix~\ref{appendix:prompts_eval_error_span_detect}).

We evaluate the predicted spans using the Intersection over Union (IoU).
A predicted span is considered a true positive if there exists a ground-truth span in our dataset such that the word-level IoU between the predicted and true span exceeds a predefined strictness threshold $t$, where $t=0$ accepts any word overlap and $t=1$ requires an exact match. 
To determine the optimal threshold, we compare this overlap-based heuristic against an LLM-driven semantic judge that evaluates based on error rationales. We find that setting $t = 0.2$ yields the highest alignment (average F1-score: 0.79; median: 0.80) between these two methods across all evaluated models (Appendix~\ref{appendix:eval_details_span_detect_setup}).

\paragraph{Erroneous sentence classification.} The second evaluation is a binary classification task that uses the sentence-level version of our dataset. Given a specific target sentence, the LLM-judge is prompted to determine whether the sentence contains any cultural or linguistic errors (True/False). To ensure that models have adequate information to resolve potential ambiguities, the target sentence is presented alongside the query and the full response (Appendix~\ref{appendix:prompts_eval_sent_classify}).

\begin{figure}
    \centering
    \includegraphics[width=\textwidth]{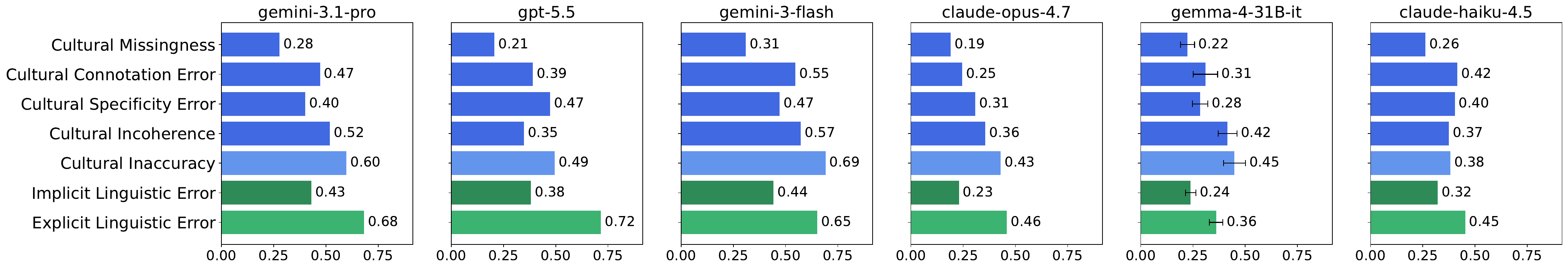}
    \caption{Category-wise recall of LLM-judges with top-6 F1 scores on the span detection task.}
    \label{fig:category_recall_span_detection}
\end{figure}

\subsection{Evaluation results}

\subsubsection{Erroneous span detection}

\autoref{tab:scores_span_detection} reports span-level precision, recall, and F1 against the validated \dataset{} annotations at an IoU threshold of 0.2 (Results across different thresholds are provided in Appendix \ref{appendix:eval_details_span_detect_result}). The best F1 in this setting is 0.51 from \texttt{gemini-3.1-pro-preview}, followed closely by \texttt{gpt-5.5} at 0.50 and \texttt{gemini-3-flash-preview} at 0.46. The remaining models score lower, including \texttt{claude-opus-4.7} at 0.44 and the open-source models at 0.25--0.43. Overall, these results show that span-level error detection remains difficult even for the best-performing judges.

To account for any potential influence stemming from the use of \texttt{gemini-3-pro} and \texttt{gpt-5.1} during dataset construction, we present an additional evaluation result strictly on the human-detected error subset (\autoref{fig:category_recall_span_detection_human} in Appendix~\ref{appendix:eval_details_span_detect_result}).
Focusing on recall, we observe that overall performance remains quite limited; \texttt{gemini-3-flash} achieves a maximum score of just 0.48, followed by \texttt{gemini-3.1-pro} at 0.46.
We note that F1 scores on this independent set fall no higher than the 0.26 achieved by \texttt{gemma-4-31B-it}, followed by \texttt{claude-opus-4.7} (0.25) and \texttt{gemini-3.1-pro-preview} (0.25).

To examine performance across error types, we report category-wise recall (\autoref{fig:category_recall_span_detection}; see \autoref{fig:category_recall_span_detection_human} for human-detected errors). Across all six top-performing models in \autoref{fig:category_recall_span_detection}, recall is highest for \textit{Explicit Linguistic Errors} (0.36--0.72) and \textit{Cultural Inaccuracy} (0.38--0.69)---the two thin categories defined by surface-level or verifiable mistakes. Recall then drops sharply for thicker cultural categories: \textit{Cultural Specificity} (0.28--0.47), \textit{Cultural Connotation} (0.25--0.55), and most strikingly \textit{Cultural Missingness}, which falls to 0.19--0.31 even for the strongest judges. The same pattern holds for human-detected errors. In other words, current LLM-judges struggle most precisely where cultural judgment matters most---on errors that require deeper cultural reasoning rather than surface-level pattern matching.
However, this decline in recall for thicker errors is less pronounced in models with lower overall detection and localization performance.

\subsubsection{Erroneous sentence classification}

We complement the span-level analysis with the sentence-level binary classification task, which removes sensitivity to span boundaries while preserving the localization of judgment (Appendix~\ref{appendix: eval_result_classification}). On this task, \texttt{gemini-3.1-pro-preview} (F1 0.57) and \texttt{gemini-3-flash-preview} (F1 0.52) again lead, followed by \texttt{gpt-5.5} (0.50) and \texttt{claude-haiku-4.5} (0.40). 
Most models achieve high accuracy (0.78--0.85), but this largely reflects the class imbalance, as clean sentences far outnumber erroneous ones. Recall is substantially lower than precision for all models except \texttt{claude-haiku-4.5} and \texttt{Qwen3-30B-A3B-Instruct-2507}.
\texttt{claude-opus-4.7} illustrates this gap most starkly: its high precision (0.77) is undercut by very low recall (0.24), yielding an F1 of 0.37 despite high accuracy of 0.85.

In terms of category-wise recall, sentence classification shows a similar trend with span detection task. Recall is highest on \textit{Explicit Linguistic Errors} (0.31--0.65) and \textit{Cultural Inaccuracy} (0.25--0.60), and lowest on \textit{Cultural Missingness} (0.10--0.31). The fact that this pattern holds across both tasks suggests that the gap is not simply a byproduct of span-boundary noise or annotation granularity, but reflects a broader difficulty in recognizing thick cultural errors.

\input{sources/minipage_improvement}

\subsection{Guiding LLM-judges with thick taxonomy}

To investigate how explicit guidance affects the model's ability to detect nuanced errors, we evaluate performance changes when providing our thick error taxonomy definition and few-shot examples. For these experiments, we employ \texttt{gemma-4-31B-it} \citep{google2026gemma4}, which demonstrated the best performance among open-source models in our previous baseline experiments.
(We additionally conduct experiments with \texttt{gemini-3.1-pro-preview} and \texttt{gemini-3-flash-preview}, with the results reported in Appendix~\ref{appendix:improvement_gemini}.)

We compare a baseline setting, in which the model is simply instructed to identify any cultural or linguistic errors, with four augmented settings. The first (\emph{+ Taxonomy Definition}) adds explicit definitions of our taxonomy categories to the prompt.
The second (\emph{+ Taxonomy Examples}) uses few-shot prompting. For each of the seven error categories, we sample two paragraphs containing at least one error of that category.
Each demonstration includes the complete list of annotated errors in the sampled paragraph, regardless of their categories.
We additionally include two error-free paragraphs as negative examples and randomly order all demonstrations.
The third (\emph{+ Taxonomy Definition + Taxonomy Examples}) combines both the taxonomy definitions and the few-shot demonstrations in a single prompt.
Finally, the fourth (\emph{+ Only Thin Examples}) serves as an ablation in which the demonstrations are restricted to the two thin error categories, \emph{Explicit Linguistic Error} and \emph{Cultural Inaccuracy}, comprising 14 paragraphs with only thin errors and two error-free ones.
We run four trials with different prompts for each setting and report the average scores. Further details are provided in Appendix \ref{appendix:improvement_setting}.

\autoref{tab:improvement} presents the overall F1 score, precision, and recall across the different settings.
The taxonomy-guided settings exhibit a consistent trade-off: a notable increase in recall accompanied by a decrease in precision, ultimately leading to a marginal improvement in the F1 score. Among the configurations tested, the \emph{+ Taxonomy Definition + Taxonomy Examples} setting yields the best overall performance.
\autoref{fig:improvement_category} further breaks down the recall improvements by error category.
The \emph{+ Taxonomy Examples} and \emph{+ Taxonomy Definition + Taxonomy Examples} settings in the Cultural Incoherence category show statistically significant improvements over the baseline (t-test, $p < 0.05$).
Notably, recall gains are greater for thick errors than for thin ones. Under the \emph{+ Taxonomy Definition + Taxonomy Examples} setting, recall increases by 4.5\%p for thin categories and 6.8\%p for thick categories.
In contrast, the \emph{+ Only Thin Examples} setting reduces recall in four of the five thick error categories. This suggests that exposure to error examples alone is insufficient and the types of errors in those examples materially affect which errors the judge can identify. More detailed results are presented in Appendix~\ref{appendix:improvement_gemma}.
Taken together, these findings demonstrate that explicit guidance is helpful for detecting errors that require deeper cultural interpretation. Nevertheless, the absolute performance on thick errors remains lower on average than that on thin errors, indicating that prompt-level guidance alone is not sufficient to close the gap.

%% file: sources/tab_scores_span_dectection.tex
\begin{table}
    \caption{Span detection performance on \dataset{}.}
    \label{tab:scores_span_detection}
    \centering
    \begin{tabular}{llll}
    \toprule
    Model & Precision & Recall & F1 score \\
    \midrule
    gemini-3.1-pro-preview & $0.4646$ & $0.5535$ & $0.5052$ \\
    gpt-5.5 & $0.4779$ & $0.5158$ & $0.4961$ \\
    \midrule
    claude-haiku-4.5 & $0.3205$ & $0.3854$ & $0.3500$ \\
    claude-opus-4.7 & $0.5509$ & $0.3617$ & $0.4367$ \\
    gemini-3-flash-preview & $0.3857$ & $0.5778$ & $0.4626$ \\
    gpt-5.4-mini & $0.3998$ & $0.2252$ & $0.2881$ \\
    \midrule
    gemma-4-31B-it & $0.5947_{\pm 0.0443}$ & $0.3401_{\pm 0.0318}$ & $0.4300_{\pm 0.0170}$ \\
    gpt-oss-120b & $0.3123_{\pm 0.0211}$ & $0.3860_{\pm 0.0196}$ & $0.3442_{\pm 0.0096}$ \\
    Llama-4-Scout-17B-16E-Instruct & $0.1979_{\pm 0.0266}$ & $0.3788_{\pm 0.0908}$ & $0.2506_{\pm 0.0136}$ \\
    Qwen3-30B-A3B-Instruct-2507 & $0.2090_{\pm 0.0112}$ & $0.4338_{\pm 0.0352}$ & $0.2810_{\pm 0.0058}$ \\
    \bottomrule
    \end{tabular}
\end{table}

%% file: sources/minipage_improvement.tex
\begin{figure}[t]
    \centering
    \begin{minipage}{0.55\linewidth}
        \centering
        \captionof{table}{Span detection performance of \texttt{gemma-4-31B-} \texttt{it} under different prompting conditions.}
        \label{tab:improvement}
        \vspace{1mm}
        {\small
        \begin{tabular}{llll}
            \toprule
            Prompt & F1 score & Precision & Recall \\
            \midrule
            Span detection & $0.4307$ & $0.5976$ & $0.3398$ \\ \midrule[0.3pt]
            + Only Thin Examples & $0.4259$ & $0.5800$ & $0.3386$ \\ \midrule[0.3pt]
            + Taxonomy Examples & $0.4345$ & $0.5210$ & $0.3792$ \\ \midrule[0.3pt]
            + Taxonomy Definition & $0.4378$ & $0.5490$ & $0.3671$ \\ \midrule[0.3pt]
            + Taxonomy Definition & \multirow{2}{*}{$0.4407$} & \multirow{2}{*}{$0.5208$} & \multirow{2}{*}{$0.3851$} \\
            + Taxonomy Examples \\
            \bottomrule
        \end{tabular}
        }
    \end{minipage}
    \hfill
    \begin{minipage}{0.4\linewidth}
        \centering
        \includegraphics[width=\linewidth]{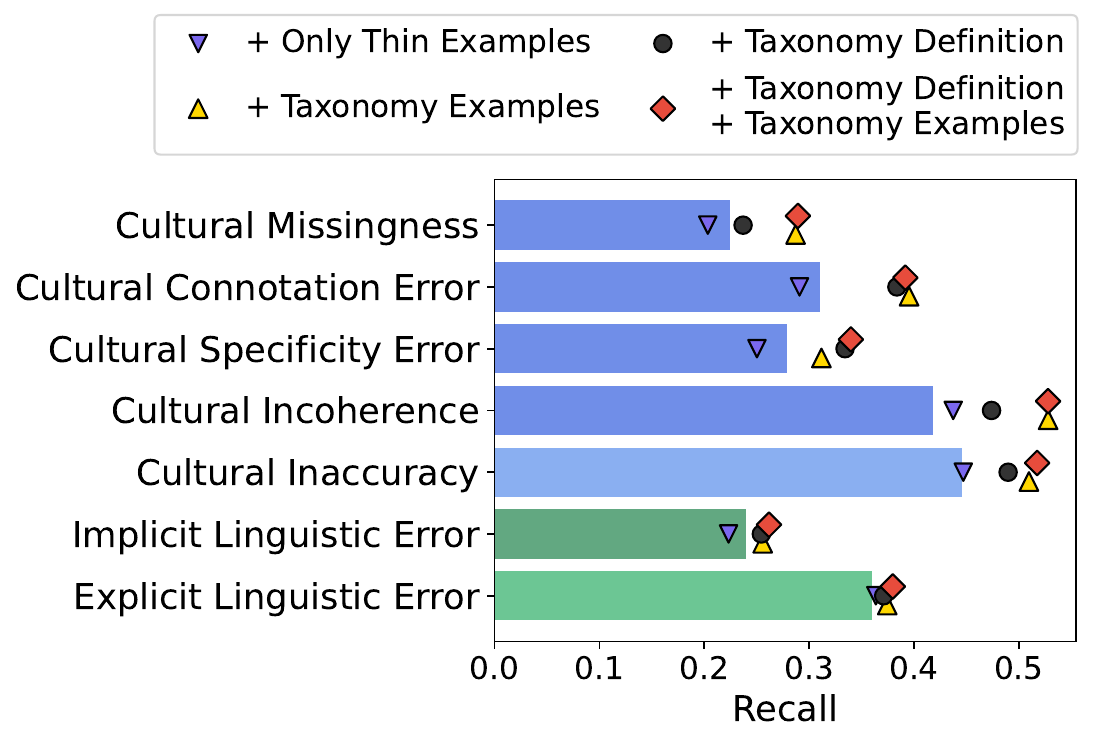}
        \captionof{figure}{Category-wise recall for \texttt{gemma-4-31B-it} under different prompting conditions.}
        \label{fig:improvement_category}
    \end{minipage}
\end{figure}

%% file: contents/05_Conclusion.tex
We introduce \dataset{}, a multilingual benchmark for evaluating whether LLM-judges can detect cultural and linguistic errors in long-form generation. Our results show that current LLM-judges are substantially better at identifying thin, surface-level errors than thick cultural errors that require situated and interpretive judgment. This gap suggests that robust cultural evaluation must move beyond factual verification toward frameworks that better capture contextual fit, connotation, and missing cultural grounding. A promising direction for future work is to develop judge models and evaluation protocols that explicitly target thick errors, for example, by incorporating culturally grounded rubrics or retrieval from local knowledge sources.

We also acknowledge several limitations of this work.
First, \dataset{} is not intended to exhaustively capture every error in each response; rather, it provides a conservative set of human-validated errors. Cultural judgments are inherently plural and subjective, and what counts as an error may vary across social groups. As a result, some subtle or debatable errors may remain unannotated, especially when they fall outside our annotators' experience. These judgments may also evolve over time as cultural norms and language use change, so our annotations reflect judgments made within a particular temporal context. In addition, because our annotator pool was recruited primarily from university-affiliated populations, the dataset may overrepresent the kinds of errors noticed by younger and more highly educated speakers. Given its limited scale and cultural coverage, \dataset{} should be viewed as an initial diagnostic benchmark, with broader recruitment across age, education, and other demographic dimensions needed to better capture diverse perspectives within each community.

Second, the distribution and difficulty of errors differ across countries and languages. In some country-language settings, responses contain more explicit or surface-level errors, while in others, the errors are more implicit or culturally profound. Thus, country-level scores should not be interpreted as direct measures of how well LLM-judges understand one culture compared to another.
Third, prompt-level interventions cannot compensate for cultural knowledge absent from a model and may offer only limited gains in detecting thick cultural errors. We therefore view taxonomy guidance and few-shot examples as aids for eliciting and applying existing knowledge, rather than substitutes for learning from culturally grounded training data.
Finally, some parts of our pipeline—including category assignment and post-processing—rely on LLM assistance. Although these steps were designed to improve consistency and scalability, the resulting labels should be understood as structured approximations of human judgment rather than as a single authoritative ground truth.

%% file: contents/X_Appendix.tex
\section{Ethical considerations}
\label{appendix:ethics}
This research project was performed under approval from KAIST IRB (KAISTIRB-2025-246).
We ensured that annotators' wages exceeded the country's minimum wage and the recommended amount from Prolific.
There was no discrimination in recruiting workers based on demographics, including gender and age.

The dataset is designed to measure LLM-Judges' capabilities for detecting thick cultural and linguistic errors in long-form LLM-generated text.
We explicitly state the terms of use in that we do not condone any malicious use.
We strongly encourage researchers and practitioners to use this dataset for beneficial purposes, such as building more culturally inclusive, globally aware AI systems that respect local customs and sociolinguistic norms.

Users of this dataset might falsely assume that the cultural norms annotated here represent an absolute, monolithic truth. To mitigate this, the dataset includes rich annotations detailing why an error was flagged.
 We also strongly recommend that users review the annotator demographic metadata to understand the specific lens through which the annotations were made.

Gemini\footnote{\url{https://gemini.google.com/}}, ChatGPT\footnote{\url{https://chatgpt.com/}}, and Claude\footnote{\url{https://claude.ai}} were used for writing and coding assistance.
Asta\footnote{\url{https://asta.allen.ai/}} was used for literature search assistance.

\section{Taxonomy}
\label{appendix:taxonomy}
\input{sources/tab_taxonomy_detail}

\section{Prompts for dataset construction}
\label{appendix:prompts_data}

\subsection{Response generation}
\label{appendix:prompts_data_response}

\begin{tcolorbox}[
breakable, enhanced, top=1pt, left=1pt, right=1pt, bottom=1pt, colback=white, fontupper=\ttfamily, fonttitle=\bfseries, 
title={System Prompt}
]
Respond in \{language\} within 400 words.

\tcbsubtitle{Prompt}

\{query\}
\end{tcolorbox}

\subsection{Grammar correction}
\label{appendix:prompts_data_grammar}
\begin{tcolorbox}[
breakable, enhanced, top=1pt, left=1pt, right=1pt, bottom=1pt, colback=white, fontupper=\ttfamily, fonttitle=\bfseries,
title={Prompt}
]
\small
\#\#\# Task Description
\\
\begin{itemize}[label={-}, itemsep=0em, leftmargin=2em]
\item Proofread the input text for grammar, spelling, and punctuation errors only.
\item Correct the grammar while keeping everything else unchanged as much as possible.
\item Output the corrected version only.
\end{itemize}\leavevmode
\\
\#\#\# Input
\\
\{text\}
\end{tcolorbox}

\subsection{Human-detected error refinement}
\label{appendix:prompts_human_refinement}
\begin{tcolorbox}[
breakable, enhanced, top=1pt, left=1pt, right=1pt, bottom=1pt, colback=white, fontupper=\ttfamily, fonttitle=\bfseries,
title={Prompt}
]
\small
Your task is to refine raw human annotations into high-quality, standardized data points for an error-detection dataset.
\\
Human annotators evaluated LLMs' responses to a specific query to detect errors.
\\
They provided a raw error span, an explanation of the error, and occasionally a suggested correction.
\\
A secondary validator reviewed these annotations and optionally provided comments. 
\\
\\
\#\#\# Objective
\\
\\
Synthesize the provided inputs and generate a final, polished `refined\_error\_span' and `refined\_explanation'.
\\
\\
\#\#\# Input Data
\\
\begin{itemize}[label={-}, itemsep=0em, leftmargin=2em]
\item **Query:** \{query\}
\item **LLM Response:** \{response\}
\item **Raw Error Span:** \{span\}
\item **Raw Explanation:** \{explanation\}
\item **Suggested Correction (Optional):** \{correction\}
\item **Validator Comment (Optional):** \{validation\_comment\}
\end{itemize}\leavevmode
\\
\#\#\# Refinement Guidelines
\\
\begin{enumerate}[itemsep=1em, leftmargin=2em]
\item **Exact Span Extraction:** The `refined\_error\_span' MUST be an exact, continuous substring extracted directly from the ``LLM Response''. Refine the boundaries of the ``Raw Error Span'' to include necessary context or remove trailing/irrelevant words, but do not modify the text itself. Ensure the span is concise; avoid including excessive surrounding text that is not directly part of the error.
\item **Explanation Clarity \& Comprehensibility:** Analyze the raw inputs to understand the underlying reason why the annotators considered this an error. Rewrite the explanation to clearly convey their rationale. Focus on the core reasoning and avoid adding extraneous information or overly detailed context that was not present in the original annotation.
\item **Language Profile (English as Matrix Language):** Write the `refined\_explanation' in English. You may also use code-switched text where English acts as the matrix (base) language, embedding specific expressions, words, or idioms from the original language of the text. Use this code-switching if retaining the original terms helps explain the error more accurately or naturally.
\item **Incorporate Validator Feedback:** If a ``Validator Comment'' exists, use it to resolve disagreements, correct the raw annotator's mistakes, or add nuanced context to your refined explanation. In cases where the Validator Comment indicates that the detected span is not actually an error, set the value of `refined\_error\_span' to `null' and provide a brief justification in the `refined\_explanation'.
\item **Leverage Corrections:** If a ``Suggested Correction'' is present, use it to strengthen the refined explanation.
\end{enumerate}\leavevmode
\\
\#\#\# Output Format
\\
\\
Provide your output strictly in the following JSON format without any markdown blocks or additional text:
\\
\{
\\
``refined\_error\_span'': ``The exact erroneous substring from the LLM Response'' or null,
\\
``refined\_explanation'': ``A clear and polished explanation of why the span is erroneous''
\\
\}
\end{tcolorbox}

\subsection{Model-detected error refinement}
\label{appendix:prompts_model_refinement}
\begin{tcolorbox}[
breakable, enhanced, top=1pt, left=1pt, right=1pt, bottom=1pt, colback=white, fontupper=\ttfamily, fonttitle=\bfseries,
title={Prompt}
]
\small
Your task is to refine raw LLM-generated annotations into high-quality, standardized data points for an error-detection dataset.
\\
An LLM initially evaluated another LLM's response to a specific query to detect errors. It provided a raw error span and an explanation of the error.
\\
Human validators then reviewed these initial annotations and optionally provided comments.
\\
If a human validator comment is present, give it priority, as it reflects human oversight over the initial LLM evaluation.
\\
\\
\#\#\# Objective
\\
\\
Synthesize the provided inputs and generate a final, polished `refined\_error\_span' and `refined\_explanation'.
\\
\\
\#\#\# Input Data
\begin{itemize}[label={-}, itemsep=0em, leftmargin=2em]
\item **Query:** \{query\}
\item **LLM Response:** \{response\}
\item **Raw Error Span:** \{span\}
\item **Raw Explanation:** \{explanation\}
\item **Validator Comment:** \{validation\_comment\}
\end{itemize} \leavevmode
\\
\#\#\# Refinement Guidelines
\\
\begin{enumerate}[itemsep=1em, leftmargin=2em]
\item **Exact Span Extraction:** The `refined\_error\_span' MUST be an exact, continuous substring extracted directly from the ``LLM Response''. Refine the boundaries of the ``Raw Error Span'' to include necessary context or remove trailing/irrelevant words, but do not modify the text itself. Ensure the span is concise; avoid including excessive surrounding text that is not directly part of the error.
\item **Prioritize Human Validator Feedback:** If a ``Validator Comment'' exists, use it as your primary guide. Incorporate human validators' insights to resolve any disagreements, correct initial mistakes, or add nuanced context to appropriately adjust the `refined\_error\_span' and `refined\_explanation'. In cases where the Validator Comment indicates that the detected span is not actually an error, set the value of `refined\_error\_span' to `null' and provide a brief justification in the `refined\_explanation'.
\item **Language Profile (English as Matrix Language):** Write the `refined\_explanation' in English. You may also use code-switched text where English acts as the matrix (base) language, embedding specific expressions, words, or idioms from the original language of the text. Use this code-switching if retaining the original terms helps explain the error more accurately or naturally.
\item **Use Raw Explanation When Appropriate:** If there is no ``Validator Comment'' and the ``Raw Explanation'' is already clear, well-written, and meets the Language Profile guidelines, use the original ``Raw Explanation'' as the `refined\_explanation' without unnecessary modifications.
\end{enumerate}\leavevmode
\\
\#\#\# Output Format
\\
\\
Provide your output strictly in the following JSON format without any markdown blocks or additional text:
\\
\{
\\
``refined\_error\_span'': ``The exact erroneous substring from the LLM Response'' or null,
\\
``refined\_explanation'': ``A clear and polished explanation of why the span is erroneous''
\\
\}
\end{tcolorbox}

\subsection{Error grouping}
\label{appendix:prompts_grouping}
\begin{tcolorbox}[
breakable, enhanced, top=1pt, left=1pt, right=1pt, bottom=1pt, colback=white, fontupper=\ttfamily, fonttitle=\bfseries,
title={Prompt}
]
\small
Your task is to analyze two error annotations and determine whether they point to the same underlying issue.
\\
\\
You will be provided with:
\begin{enumerate}[itemsep=0em, leftmargin=2em]
\item **Context**: The original text where the errors were found.
\item **Error A**: Contains a text span and an explanation.
\item **Error B**: Contains a text span and an explanation.
\end{enumerate}\leavevmode
\\
Guidelines:
\begin{enumerate}[itemsep=0em, leftmargin=2em]
\item **Same Underlying Error**: The errors stem from the same root cause or fundamental flaw.
    \begin{itemize}[label={-}, itemsep=0em, leftmargin=2em]
    \item The text spans do not need to be perfectly identical. They may overlap or encompass different words, as long as they target the same localized issue.
    \item The explanations do not need to be phrased identically. Focus on the core intent and reasoning of the annotation.
    \end{itemize}
\item **Different Errors**: The errors refer to distinct issues.
    \begin{itemize}[label={-}, itemsep=0em, leftmargin=2em]
    \item Even if the text spans are identical, if the explanations point out completely different problems, then they should be considered as different.
    \end{itemize}
\end{enumerate}\leavevmode
\\
\#\#\# Input Data
\\
\\
**Context:**
\begin{itemize}[label={ }, itemsep=0em, leftmargin=2em]
\item \{query\}
\item \{response\}
\end{itemize}\leavevmode
\\
**Error A:**
\begin{itemize}[label={-}, itemsep=0em, leftmargin=2em]
\item **Erroneous Span:** ``\{span\_a\}''
\item **Explanation:** ``\{explanation\_a\}''
\end{itemize}\leavevmode
\\
**Error B:**
\begin{itemize}[label={-}, itemsep=0em, leftmargin=2em]
\item **Erroneous Span:** ``\{span\_b\}''
\item **Explanation:** ``\{explanation\_b\}''
\end{itemize}\leavevmode
\\
\\
\#\#\# Output Format
\\
\\
``same'' or ``different''
\end{tcolorbox}

\subsection{Category classification}
\label{appendix:prompts_data_category_classification}
\begin{tcolorbox}[
breakable, enhanced, top=1pt, left=1pt, right=1pt, bottom=1pt, colback=white, fontupper=\ttfamily, fonttitle=\bfseries,
title={Prompt}
]
\small
\# Task Description
\\
\\
You are a classifier that categorizes a specific error identified within a model-generated response.
\\
As input, you will receive:
\begin{itemize}[label={-}, itemsep=0em, leftmargin=2em]
\item A query,
\item A response generated by a model,
\item A text span identified as containing an error, and
\item An explanation describing why it is an error.
\end{itemize}\leavevmode
\\
You are provided with a set of Error Categories.
\\
Your task is to:
\begin{itemize}[label={-}, itemsep=0em, leftmargin=2em]
\item Analyze the provided error span and explanation from the perspective of the country specified in the query.
\item Determine which specific Error Category best fits the error.
\item Output only the name of the selected error category.
\end{itemize}\leavevmode
\\
\\
\#\# Error Categories
\begin{itemize}[label={-}, itemsep=1em, leftmargin=2em]
\item **Explicit Linguistic Error**: Surface-level, mechanical language mistakes. This covers objective violations of language rules, including incorrect grammar, spelling errors, typos, literal translations that misuse vocabulary, or outputting text in the wrong language. These flaws are independent of social context.
\item **Implicit Linguistic Error**: Failures in tone, register, or sociolinguistic nuance. The text may be grammatically correct but reads unnaturally (e.g., ``awkward phrasing''), or completely misses the unspoken communication rules of the target language.
\item **Cultural Inaccuracy**: Objectively incorrect verifiable facts about a culture. Classify any factual errors regarding a specific country's artifacts, geography, local ingredients, traditions, historical events, institutions, or demographic makeup here.
\item **Cultural Incoherence**: A contextual clash of cultural elements. The cultural elements mentioned are real, but combining them in the given setting, situation, or alongside other elements is jarring, awkward, or highly unnatural.
\item **Cultural Specificity Error**: Misrepresenting the boundaries of a cultural concept. This occurs when the model over-generalizes (applying a specific trait to an entire population via explicit bias/stereotyping), under-generalizes (treating a broad cultural norm as overly narrow), or conflates distinct regional variations into a single monolith.
\item **Cultural Connotation Error**: Mishandling the implicit symbolic weight or emotional resonance of a concept. This applies when the response fails to grasp deep-seated symbolic, spiritual, or sentimental meanings (e.g., treating a sacred object as a trivial commodity), or carries unintended negative baggage through subtle framing and word choice.
\item **Cultural Missingness**: The absence of critical, non-negotiable cultural context required for the response to be functionally or contextually complete. This applies only when the omission of specific traditions, norms, or nuances results in a ``hollow'' or ``generic'' output that fails to address the unique cultural identity of the subject.
\item **Logical Error**: Pure reasoning failures independent of language or culture. This includes internal contradictions, invalid deductions, or scientific/mathematical flaws that have nothing to do with cultural context or linguistic mechanics.
\end{itemize}\leavevmode
\\
\\
\# Examples
\begin{itemize}[label={-}, itemsep=1em, leftmargin=2em]
\item Explicit Linguistic Error:
\\
Error Span: fresh soy milk (kacang kedelai)
Explanation: In Indonesian, the common and correct term for soy milk is ``susu kedelai,'' not ``kacang kedelai.'' ``Kacang kedelai'' refers to the soybean itself (the bean), not the milk made from it, so this is a linguistic misuse of the phrase.
\\
Error Span: hwangap 60
Explanation: “Hwangap” by definition refers to the 60th birthday, so adding “60” after it is repetitive and unidiomatic. It would be clearer to say “hwangap (60th birthday)” or simply “hwangap.”
\item Implicit Linguistic Error:
\\
Error Span: Nailed It!
Explanation: The English idiom 'Nailed It!' is colloquial and might not be immediately understood or carry the intended encouraging weight for a Korean student audience, even if they study English. A culturally localized product would likely use Korean phrases like 'Hwaiting!' (Fighting/Cheer up) or 'Hapgyeok' (Pass/Success) specifically relevant to exam culture.
\\
Error Span: 포용: 휠체어 학생은 패스하거나 별도 점수 방식
Explanation: The term '패스하다' (to pass or skip) is culturally insensitive and linguistically awkward, particularly under the heading '포용' (inclusion). In a Korean educational context, suggesting that students in wheelchairs simply 'pass' contradicts the core concept of inclusive practices. A more appropriate approach would involve assigning alternative roles, such as '심판이나 코치 역할 부여' (assigning a referee or coach role) or '대체 역할 수행' (performing an alternative role). Furthermore, '패스하다' is informal Konglish in this context; while '참여 제외' or '열외' are standard terms for exclusion, they are inherently negative and unsuitable for a section dedicated to inclusion.
\item Cultural Inaccuracy:
\\
Error Span: environmental science
Explanation: Universitas Gadjah Mada (UGM) does not offer an undergraduate bachelor's degree program specifically in Environmental Science. The closest equivalent major available at the university is Environmental Geography.
\\
Error Span: commanding buses with a shout
Explanation: From a cultural perspective, a city bus driver (Darnell) does not have the authority to step into the road and start directing or commanding other buses or traffic simply by shouting, especially in the presence of a traffic cop (Officer Rao). This action feels unrealistic for the role of a bus driver in a standard US city context.
\item Cultural Incoherence:
\\
Error Span: making traditional snacks (e.g., *klepon* or *lumpia*)
Explanation: Making traditional snacks like klepon or lumpia feels out of place in an educational garden setting. It does not align with the other listed activities, which are strictly focused on farming and gardening.
\\
Error Span: isian sayur (capcay)
Explanation: Capcay is a Chinese-Indonesian stir-fried vegetable dish that typically has a watery or thick sauce. Using 'capcay' as a filling for a rolled pizza or pizza cone is culinarily impractical, as the wet consistency of the dish would make the pizza dough soggy and messy to eat. This contradicts the premise of the snack being practical and easy to eat while walking around an amusement park.
\item Cultural Specificity:
\\
Error Span: kegiatan keagamaan/pengajian
Explanation: The use of the slash in 'kegiatan keagamaan/pengajian' incorrectly implies that 'kegiatan keagamaan' (religious activities) and 'pengajian' (Islamic recitation/study) are synonymous. 'Kegiatan keagamaan' is a broader term, and since Indonesia officially recognizes six religions, equating it specifically with an Islamic practice inaccurately assumes that the students are Muslim.
\\
Error Span: modest dress only required at religious sites.
Explanation: The claim that modest dress is exclusively required at religious sites is inaccurate. Due to cultural norms in Indonesia, modest attire is generally expected and appreciated in most public areas, including malls, markets, and formal events.
\item Cultural Connotation Error:
\\
Error Span: The scent of jasmine, typically a harbinger of peace
Explanation: In Indonesian culture, especially Javanese (implied by 'Kakek' and 'pendopo'), the strong scent of jasmine ('melati') is rarely associated simply with 'peace'. Instead, it is heavily associated with mysticism, spirits, death, and the supernatural. While jasmine flowers are used in funeral rituals (like washing the body and scattering flowers), the sudden smell of jasmine is culturally interpreted as a sign of a spirit's presence, often creating a sense of spookiness or reverence rather than peacefulness.
\\
Error Span: A Black man in his late forties, towering over teenagers with a voice roughened by years of shouting over factory noise, didn’t seem like the obvious choice for a tutor.
Explanation: This sentence can be read as implying that being a Black man with a manual-labor background makes someone an unlikely or unsuitable tutor, which risks reinforcing negative racial and class stereotypes. Although it may be intended to reflect others’ biased perceptions, the narrative does not clearly mark it as such, so it can be culturally insensitive.
\item Cultural Missingness:
\\
Error Span: umat Hindu, Kristen, Buddha, dan kepercayaan lokal.
Explanation: The LLM lists some of the minority religions in Indonesia but omits Confucianism (Konghucu) and Catholicism (Katolik). Since the Indonesian government officially acknowledges six religions (Islam, Protestantism/Kristen, Catholicism, Hinduism, Buddhism, and Confucianism), it is important to include all of them for accuracy.
\\
Error Span: soda
Explanation: While drinking soda in a sports bar is possible, in the context of South Korea's 'chimaek' (chicken and beer) culture, drinking beer (maekju) is overwhelmingly the cultural norm during major football matches. Specifically gripping a 'soda' feels slightly out of place for an intense sports bar atmosphere in Seoul.
\item Logical Error:
\\
Error Span: watch sticker
Explanation: During an exam, candidates usually bring a wristwatch to keep time, not a sticker of a watch, which would be functionally useless for time management. This is likely a hallucination or a linguistic error confusing a functional 'watch' (timepiece) with a decorative item.
\\
Error Span: a slow‑moving truck—humid, heavy, and utterly unavoidable.
Explanation: The simile comparing a Monday morning to 'a slow-moving truck' is logically flawed in this context. A truck cannot be described as 'humid,' and a 'slow-moving' truck is generally avoidable, which contradicts the description that it is 'utterly unavoidable.'
\end{itemize} \leavevmode
\\
\\
\#\# Query
\\
\{query\}
\\
\\
\#\# Response
\\
\{response\}
\\
\\
\#\# Error Span
\\
\{span\}
\\
\\
\#\# Explanation
\\
\{explanation\}
\end{tcolorbox}

\FloatBarrier

\section{Annotation details}
\label{appendix:annotation}

\subsection{Annotator demographics}
\label{appendix:annotation_demographics}
\input{sources/tab_annotator_stats}
We recruited 44 native speakers across the four target countries (10 each for the US, KR, and ID, and 14 for BD). To be eligible, participants had to have spent at least half of their life in the target country and demonstrate proficiency in both English and the country's primary local language. Detailed annotator demographics is specified in Table~\ref{tab:annotator_demographics}.

\subsection{Error annotation and validation interface}
\label{appendix:annotation_interface}
\begin{figure}[ht]
    \centering
    \includegraphics[width=0.4\linewidth]{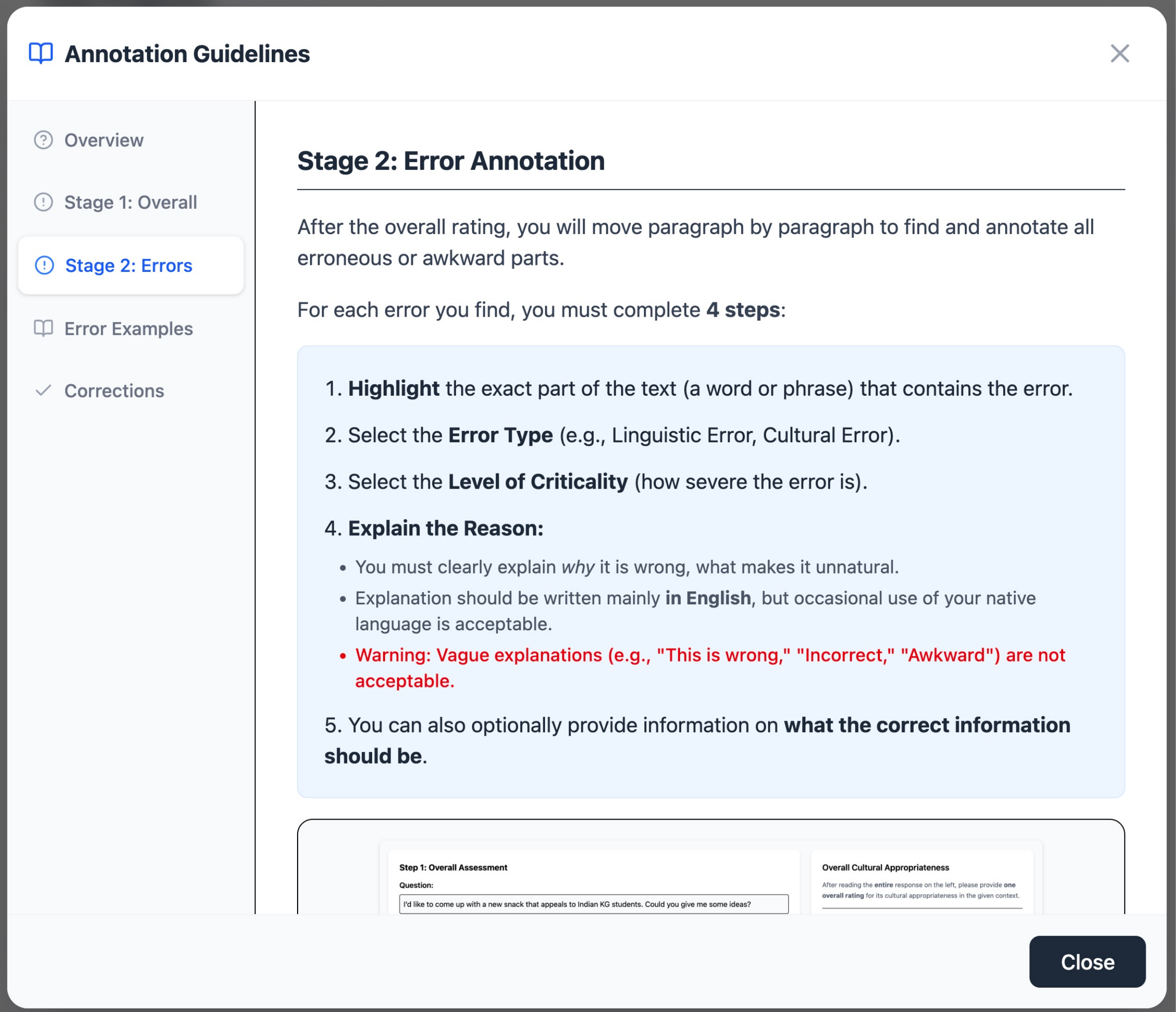}
    \caption{Annotation interface: Error annotation guideline.}
    \label{fig:annotation_interface1}
\end{figure}

\begin{figure}[ht]
    \centering
    \includegraphics[width=\linewidth]{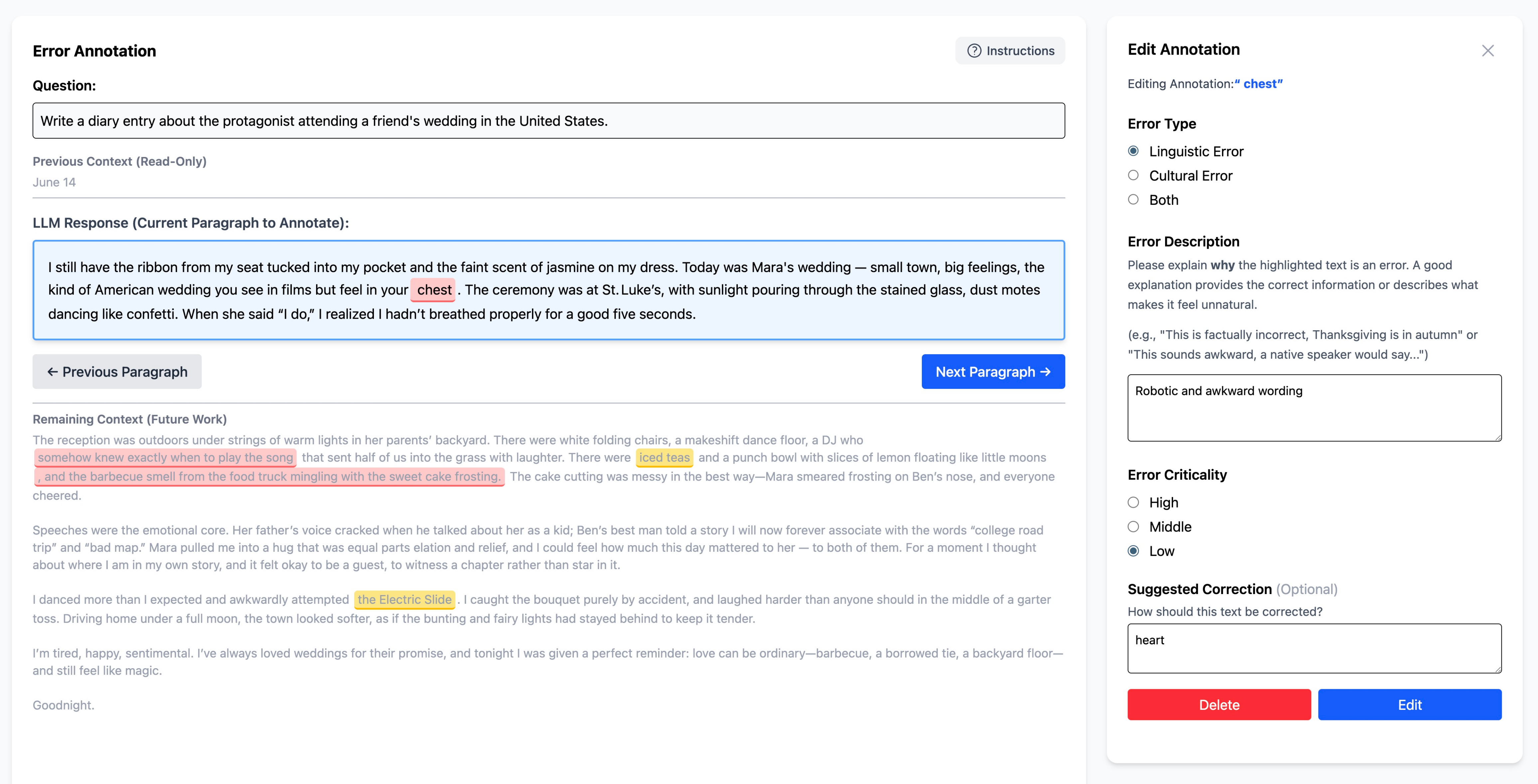}
    \caption{Annotation interface: Error annotation.}
    \label{fig:annotation_interface2}
\end{figure}

\begin{figure}[ht]
    \centering
    \includegraphics[width=\linewidth]{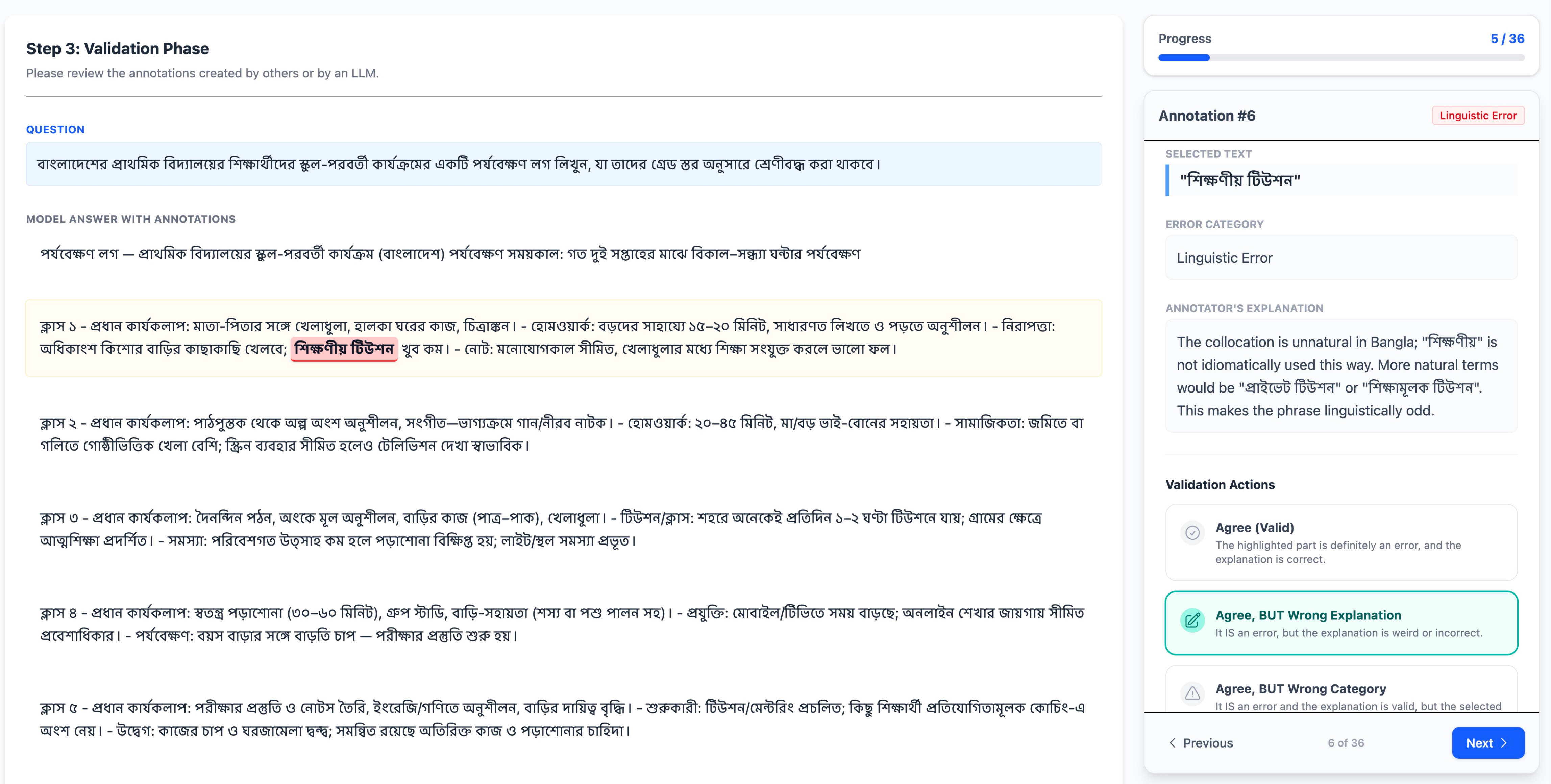}
    \caption{Annotation interface: Error validation.}
    \label{fig:annotation_interface3}
\end{figure}

Figure~\ref{fig:annotation_interface1} shows the annotation guidelines provided to human annotators, Figure~\ref{fig:annotation_interface2} presents the interface used for error annotation, and Figure~\ref{fig:annotation_interface3} for error validation.

\FloatBarrier

\subsection{Annotation refinement}
\label{appendix:annotation_refinement}
Raw annotations frequently contained fragmented spans, inconsistent levels of detail, or explanations written in different languages. To standardize annotations for analysis, we used \texttt{gemini-3.1-pro} to rewrite explanations into a consistent English format while incorporating validator feedback from the second annotation stage. When validator feedback conflicted with the original annotation, human-provided corrections were prioritized.

We manually reviewed 200 refined annotations to assess whether the refinement process altered annotation meaning. Only 2 annotations introduced additional details beyond the original annotation, and none introduced factual inaccuracies. The full refinement prompt is shown in Appendix~\ref{appendix:prompts_human_refinement}-\ref{appendix:prompts_model_refinement}.

\subsection{Aggregation into error groups}
\label{appendix:annotation_aggregation}
Because multiple annotators often identified the same underlying issue using partially overlapping spans, we aggregated redundant annotations into a unified error. Candidate pairs were first generated using span overlap within each response. We then prompted \texttt{gemini-3.1-pro} to determine whether paired annotations referred to the same underlying error based on both the highlighted spans and accompanying explanations.

Annotations with overlapping spans but distinct semantic concerns were retained as separate errors. We manually reviewed 50 candidate pairs to validate the quality of the aggregation, and all were valid. Full prompts used from grouping are provided in Appendix~\ref{appendix:prompts_grouping}.

\subsection{Annotation statistics}
\label{appendix:annotation_stats}

Detailed annotation statistics for each stage are specified in \autoref{tab:annotation_stats}. 
Overall, the annotation pipeline produced 7,470 validated error spans across 1,050 query-response pairs, with the number and composition of errors varying across country-language settings. 
This variation reflects differences not only in the frequency of errors, but also in the kinds of errors that annotators and model judges identified.

\begin{figure}[ht]
    \centering
    \includegraphics[width=\linewidth]{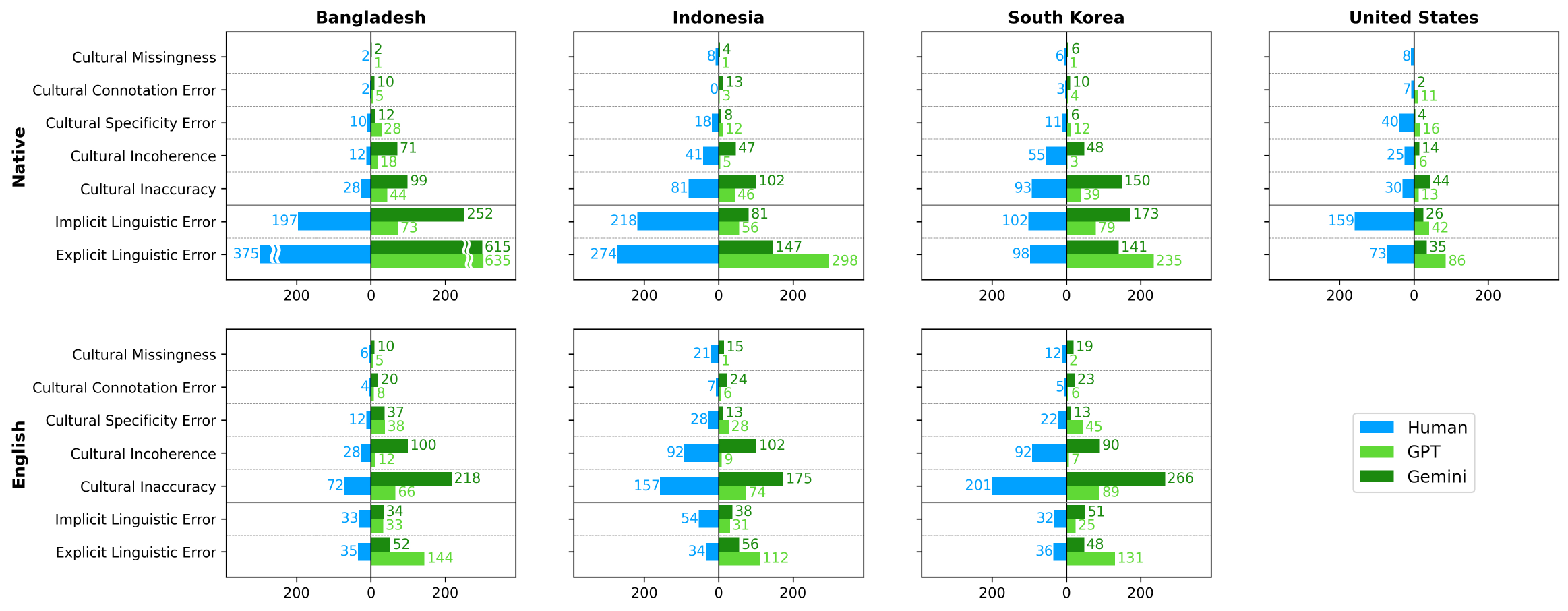}
    \caption{Distribution of annotation per class-country.}
    \label{fig:tick_dist_model_separate}
\end{figure}

\autoref{fig:tick_dist_model_separate} further breaks down the distribution of error categories by detector type and language setting. 
A clear distinction emerges between native-language and English settings. 
In native-language responses, linguistic errors constitute a much larger share of the annotations, particularly explicit linguistic errors in Bengali, Indonesian, and Korean. 
This suggests that when responses are generated in the local language, annotators are more likely to notice surface-level fluency issues as well as sociolinguistic mismatches in tone, register, or naturalness. 
By contrast, in English settings, the relative share of cultural errors increases, especially cultural inaccuracy and cultural incoherence. 
This pattern indicates that English responses often avoid some language-specific fluency problems, but still fail to capture locally appropriate cultural details.

The distribution also differs between human annotators and LLM-judges. 
Human annotators identify a broader range of implicit linguistic and culturally situated errors, whereas model judges tend to concentrate more heavily on explicit linguistic errors and cultural inaccuracies. 
This supports our main finding that LLM-judges are more sensitive to surface-level or verifiable errors than to thicker errors requiring local cultural interpretation.

\input{sources/tab_annotation_stats}

\FloatBarrier

\section{Prompts for LLM-judge evaluation}
\label{appendix:prompts_eval}

For each setting, we construct four distinct versions of the prompt. We present one representative version here, while the complete set of prompts is available in our code repository.

\subsection{Erroneous span detection}
\label{appendix:prompts_eval_error_span_detect}
\begin{tcolorbox}[
breakable, enhanced, top=1pt, left=1pt, right=1pt, bottom=1pt, colback=white, fontupper=\ttfamily, fonttitle=\bfseries,
title={Prompt}
]
\small
\#\# Task Description
\\
\\
The goal of this task is to carefully examine a given question and the corresponding response and identify any **cultural or linguistic errors** within each line in the response.
\\
\\
Given a **target line** from the response, your task is to detect any erroneous parts within it and explain the reason behind the errors.
\\
\\
Output the result strictly as a JSON array.
\begin{itemize}[label={-}, itemsep=0em, leftmargin=2em]
\item Each detected error must correspond to a valid, standalone JSON object within the array.
\item Each object must contain exactly these three keys:
    \begin{enumerate}
    \item ``span'': The specific text segment corresponding to the erroneous part (string).
    \item ``reason'': An explanation of why this is an error (string).
    \item ``error\_type'': Either ``cultural error'', ``linguistic error'', or ``both''.
    \end{enumerate}
\item If multiple errors are found, return an array containing all identified error objects.
\item If no errors are found, return an empty array ([]).
\item Do not use Markdown code blocks (e.g., \verb|```|json).
\end{itemize}\leavevmode
\\
\\
\#\# Question
\\
\{question\}
\\
\\
\\
\#\# Response
\\
\{response\}
\\
\\
\\
\#\# Target Line
\\
\{paragraph\}
\end{tcolorbox}

\subsection{Erroneous sentence classification}
\label{appendix:prompts_eval_sent_classify}
\begin{tcolorbox}[
breakable, enhanced, top=1pt, left=1pt, right=1pt, bottom=1pt, colback=white, fontupper=\ttfamily, fonttitle=\bfseries,
title={Prompt}
]
\small
\#\# Task Description
\\
\\
The goal of this task is to carefully examine the [Target Sentence] within the context of the [Query] and [Full Response] to determine if it contains any **cultural or linguistic errors**.
\\
\\
Infer the target culture from the [Query], and evaluate the [Target Sentence] from the perspective of **that specific culture**.
\\
\\
Return ``True'' if any cultural or linguistic error exists in the [Target Sentence], otherwise ``False''. Do not generate any other text.
\\
\\
\\
\#\# Query
\\
\{query\}
\\
\\
\\
\#\# Full Response
\\
\{response\}
\\
\\
\\
\#\# Target Sentence
\\
\{sentence\}
\end{tcolorbox}

\subsection{Erroneous span detection + taxonomy}
\label{appendix:improvement_tax}
\begin{tcolorbox}[
breakable, enhanced, top=1pt, left=1pt, right=1pt, bottom=1pt, colback=white, fontupper=\ttfamily, fonttitle=\bfseries,
title={Prompt}
]
\small
\#\# Task Description
\\
\\
The goal of this task is to carefully examine a given question and the corresponding response and identify any **cultural or linguistic errors** within each line in the response, based on the provided Error Taxonomy.
\\
\\
Given a **target line** from the response, your task is to detect any erroneous parts within it, classify the error according to the taxonomy, and explain the reason behind the errors.
\\
\\
\\
\#\# Error Taxonomy
\\
\begin{itemize}[label={-}, itemsep=1em, leftmargin=2em]
\item **Explicit Linguistic Error**: Surface-level, mechanical language mistakes. This covers objective violations of language rules, including incorrect grammar, spelling errors, typos, literal translations that misuse vocabulary, or outputting text in the wrong language. These flaws are independent of social context.
\item **Implicit Linguistic Error**: Failures in tone, register, or sociolinguistic nuance. The text may be grammatically correct but reads unnaturally (e.g., ``awkward phrasing''), or completely misses the unspoken communication rules of the target language.
\item **Cultural Inaccuracy**: Objectively incorrect verifiable facts about a culture. Classify any factual errors regarding a specific country's artifacts, geography, local ingredients, traditions, historical events, institutions, or demographic makeup here.
\item **Cultural Incoherence**: A contextual clash of cultural elements. The cultural elements mentioned are real, but combining them in the given setting, situation, or alongside other elements is jarring, awkward, or highly unnatural.
\item **Cultural Specificity Error**: Misrepresenting the boundaries of a cultural concept. This occurs when the model over-generalizes (applying a specific trait to an entire population via explicit bias/stereotyping), under-generalizes (treating a broad cultural norm as overly narrow), or conflates distinct regional variations into a single monolith.
\item **Cultural Connotation Error**: Mishandling the implicit symbolic weight or emotional resonance of a concept. This applies when the response fails to grasp deep-seated symbolic, spiritual, or sentimental meanings (e.g., treating a sacred object as a trivial commodity), or carries unintended negative baggage through subtle framing and word choice.
\item **Cultural Missingness**: The absence of critical, non-negotiable cultural context required for the response to be functionally or contextually complete. This applies only when the omission of specific traditions, norms, or nuances results in a ``hollow'' or ``generic'' output that fails to address the unique cultural identity of the subject.
\end{itemize}\leavevmode
\\
\\
\#\# Output Format
\\
\\
Output the result strictly as a JSON array.
\begin{itemize}[label={-}, itemsep=0em, leftmargin=2em]
\item Each detected error must correspond to a valid, standalone JSON object within the array.
\item Each object must contain exactly these three keys:
    \begin{enumerate}
    \item ``span'': The specific text segment corresponding to the erroneous part (string).
    \item ``reason'': An explanation of why this is an error (string).
    \item ``error\_type'': Must be exactly one of the seven categories listed in the Error Taxonomy above.
    \end{enumerate}
\item If multiple errors are found, return an array containing all identified error objects.
\item If no errors are found, return an empty array ([]).
\item Do not use Markdown code blocks (e.g., \verb|```|json).
\end{itemize}\leavevmode
\\
\\
\#\# Question
\\
\{query\}
\\
\\
\\
\#\# Response
\\
\{response\}
\\
\\
\\
\#\# Target Line
\\
\{paragraph\}
\end{tcolorbox}

\subsection{Erroneous span detection + few-shots}
\label{appendix:improvement_fewshots}
\begin{tcolorbox}[
breakable, enhanced, top=1pt, left=1pt, right=1pt, bottom=1pt, colback=white, fontupper=\ttfamily, fonttitle=\bfseries,
title={Prompt}
]
\small
\# Task Description
\\
\\
The goal of this task is to carefully examine a given question and the corresponding response and identify any **cultural or linguistic errors** within each line in the response.
\\
\\
Given a **target line** from the response, your task is to detect any erroneous parts within it, classify the error according to the taxonomy, and explain the reason behind the errors.
\\
\\
\\
\#\# Output Format
\\
\\
Output the result strictly as a JSON array.
\begin{itemize}[label={-}, itemsep=0em, leftmargin=2em]
\item Each detected error must correspond to a valid, standalone JSON object within the array.
\item Each object must contain exactly these three keys:
\begin{enumerate}
\item``span'': The specific text segment corresponding to the erroneous part (string).
\item``reason'': An explanation of why this is an error (string).
\item``error\_type'': Either ``cultural error'', ``linguistic error'', or ``both''.
\end{enumerate}
\item If multiple errors are found, return an array containing all identified error objects.
\item If no errors are found, return an empty array ([]).
\item Do not use Markdown code blocks (e.g., \verb|```|json).
\end{itemize}\leavevmode
\\
\\
\# Examples
\\
\\
> Example 1
\\
Question: \{query\_1\}
\\
Target Line: \{paragraph\_1\}
\\
Answer:
\{error\_list\_1\}
\\
\\
> Example 2
\\
Question: \{query\_2\}
\\
Target Line: \{paragraph\_2\}
\\
Answer:
\{error\_list\_2\}
\\
\\
...
\\
\\
\\
\# Input
\\
\\
\#\# Question
\\
\{query\}
\\
\\
\#\# Response
\\
\{response\}
\\
\\
\#\# Target Line
\\
\{paragraph\}
\end{tcolorbox}

\subsection{Erroneous span detection + taxonomy + few-shots}
\label{appendix:improvement_tax_shots}
\begin{tcolorbox}[
breakable, enhanced, top=1pt, left=1pt, right=1pt, bottom=1pt, colback=white, fontupper=\ttfamily, fonttitle=\bfseries,
title=\{Prompt\}
]
\small
\# Task Description
\\
\\
The goal of this task is to carefully examine a given question and the corresponding response and identify any **cultural or linguistic errors** within each line in the response, based on the provided Error Taxonomy.
\\
\\
Given a **target line** from the response, your task is to detect any erroneous parts within it, classify the error according to the taxonomy, and explain the reason behind the errors.
\\
\\
\\
\#\# Error Taxonomy
\\
\begin{itemize}[label={-}, itemsep=1em, leftmargin=2em]
\item **Explicit Linguistic Error**: Surface-level, mechanical language mistakes. This covers objective violations of language rules, including incorrect grammar, spelling errors, typos, literal translations that misuse vocabulary, or outputting text in the wrong language. These flaws are independent of social context.
\item **Implicit Linguistic Error**: Failures in tone, register, or sociolinguistic nuance. The text may be grammatically correct but reads unnaturally (e.g., ``awkward phrasing''), or completely misses the unspoken communication rules of the target language.
\item **Cultural Inaccuracy**: Objectively incorrect verifiable facts about a culture. Classify any factual errors regarding a specific country's artifacts, geography, local ingredients, traditions, historical events, institutions, or demographic makeup here.
\item **Cultural Incoherence**: A contextual clash of cultural elements. The cultural elements mentioned are real, but combining them in the given setting, situation, or alongside other elements is jarring, awkward, or highly unnatural.
\item **Cultural Specificity Error**: Misrepresenting the boundaries of a cultural concept. This occurs when the model over-generalizes (applying a specific trait to an entire population via explicit bias/stereotyping), under-generalizes (treating a broad cultural norm as overly narrow), or conflates distinct regional variations into a single monolith.
\item **Cultural Connotation Error**: Mishandling the implicit symbolic weight or emotional resonance of a concept. This applies when the response fails to grasp deep-seated symbolic, spiritual, or sentimental meanings (e.g., treating a sacred object as a trivial commodity), or carries unintended negative baggage through subtle framing and word choice.
\item **Cultural Missingness**: The absence of critical, non-negotiable cultural context required for the response to be functionally or contextually complete. This applies only when the omission of specific traditions, norms, or nuances results in a ``hollow'' or ``generic'' output that fails to address the unique cultural identity of the subject.
\end{itemize}\leavevmode
\\
\\
\#\# Output Format
\\
\\
Output the result strictly as a JSON array.
\begin{itemize}[label={-}, itemsep=0em, leftmargin=2em]
\item Each detected error must correspond to a valid, standalone JSON object within the array.
\item Each object must contain exactly these three keys:
    \begin{enumerate}
    \item ``span'': The specific text segment corresponding to the erroneous part (string).
    \item ``reason'': An explanation of why this is an error (string).
    \item ``error\_type'': Must be exactly one of the seven categories listed in the Error Taxonomy above.
    \end{enumerate}
\item If multiple errors are found, return an array containing all identified error objects.
\item If no errors are found, return an empty array ([]).
\item Do not use Markdown code blocks (e.g., \verb|```|json).
\end{itemize}\leavevmode
\\
\\
\# Examples
\\
\\
> Example 1
\\
Question: \{query\_1\}
\\
Target Line: \{paragraph\_1\}
\\
Answer:
\{error\_list\_1\}
\\
\\
> Example 2
\\
Question: \{query\_2\}
\\
Target Line: \{paragraph\_2\}
\\
Answer:
\{error\_list\_2\}
\\
\\
...
\\
\\
\\
\# Input
\\
\\
\#\# Question
\\
\{query\}
\\
\\
\#\# Response
\\
\{response\}
\\
\\
\#\# Target Line
\\
\{paragraph\}
\end{tcolorbox}

\FloatBarrier

\section{Evaluation details}
\label{appendix:eval_details}

\subsection{Comparison with existing cultural evaluation pipelines}
\label{appendix:eval_baseline}

To contextualize our LLM-judge results against prior cultural evaluation methodologies, we adapt two representative pipelines to our setting: a norm-entailment approach inspired by CultureBank~\citep{shi-etal-2024-culturebank} and a claim-verification approach inspired by CalmQA~\citep{arora-etal-2025-calmqa}.

\textit{Norm entailment.} Following CultureBank's evaluation pipeline, we assess whether the response entails the cultural descriptor from which the query was derived. We apply this evaluation to the 210 query-response pairs in \dataset{} sourced from CultureBank, using \texttt{gemini-3.1-pro-preview} as the LLM-judge (prompt adopted from \citet{arora-etal-2025-calmqa}).

\textit{Cultural claim verification.} CalmQA employs VeriScore~\citep{song-etal-2024-veriscore} to extract verifiable atomic claims from model outputs and verify them against external evidence. Because our responses are long-form narratives rather than factual answers, naive atomic-claim extraction yields a large fraction of culturally irrelevant claims. We therefore design a \textit{cultural claim extraction} step that prompts the model with few-shot examples to extract only the cultural facts or assumptions a sentence implies about the target country, together with the triggering span. For instance, the sentence \textit{``She hurried downstairs where her mother had made paratha and lentil soup''} yields claims such as \textit{``Paratha is a common Bangladeshi breakfast food''} (from \textit{paratha}), \textit{``Lentil soup is a common Bangladeshi breakfast food''} (from \textit{lentil soup}), and \textit{``Multi-story houses are common in Bangladesh''} (from \textit{hurried downstairs}). Each claim is then verified against two evidence sources: a curated set of Wikipedia articles related to the target country (e.g., for Bangladesh: \textit{Bangladesh}, \textit{Culture of Bangladesh}, \textit{Bangladeshi cuisine}, and similar topical pages), and Google Search results retrieved by querying the claim verbatim through the Serper API, following CalmQA. An LLM-judge then determines whether the retrieved evidence supports each claim, and any unsupported claim is treated as an error span.

We compared \texttt{gemini-3.1-pro-preview} and \texttt{gemma-4-31B-it} as cultural-claim extractors. Treating the former as the reference, \texttt{gemma-4-31B-it} semantically recovers 78.9\% of its claims (cosine $\geq$ 0.7 with \texttt{all-MiniLM-L6-v2}). Given this agreement and the substantial cost difference, we use \texttt{gemma-4-31B-it} for both claim extraction and evidence-based verification in all reported results. Full prompts for extraction are as follows. (Verification prompt adopted from VeriScore\citep{song-etal-2024-veriscore}).

\begin{tcolorbox}[
breakable, enhanced, top=1pt, left=1pt, right=1pt, bottom=1pt, colback=white, fontupper=\ttfamily, fonttitle=\bfseries,
title={Prompt}
]
\small
You are evaluating a response generated about \{COUNTRY\} culture.
\\
\\
The response may be a story, diary entry, recommendation, business idea, or travel advice.
For each sentence, extract cultural facts or assumptions it implies about \{COUNTRY\} or its people — things that reflect real cultural practices, norms, or knowledge.
\\
\\
For each claim, also quote the minimal span from the sentence that grounds it.
\\
\\
Format: - [span] cultural claim
\\
\\
If the sentence contains no verifiable cultural information (e.g. filler, generic statements, or content with no \{COUNTRY\}-specific cultural content), output exactly: N/A
\\
\\
Sentence: She hurried downstairs where her mother had made paratha and lentil soup.
\\
Cultural claims about \{COUNTRY\}:
\\
- ("partha", "Paratha is a common Bangladeshi breakfast food."),
\\
- ("lentil soup", "Lentil soup is a common Bangladeshi breakfast food.")
\\
- ("hurried downstairs", "Multi-story houses are common in Bangladesh.")
\\
\\
(5 shots specific to each country)
\\
\\
Sentence: \{TARGET SENTENCE\}
\end{tcolorbox}

\paragraph{Norm entailment underestimates errors in entailing responses.}
We first apply CultureBank's norm-entailment protocol to the 210 query--response pairs in \dataset{} sourced from CultureBank. Of these, 98 responses are judged to entail the underlying cultural descriptor, and 112 are judged not to entail it. As expected, non-entailing responses contain more error spans than entailing ones (avg.~8.63 vs.~6.08 per response; Mann--Whitney $U=4558$, $p<0.05$), and this direction holds across most culture-language pairs (with the exception of Korean-English and US-English). However, the effect size is small, and entailing responses still contain on average 6.08 error spans---meaning that even responses that pass norm-level verification contain a substantial volume of culturally problematic content that \dataset{} surfaces.

This pattern reflects a structural limitation: norm entailment verifies a single, abstract proposition while remaining insensitive to local cultural detail. For instance, in response to a query about Korean desserts, a model states that \textit{``black garlic is sometimes used in chocolate truffles or ice cream in modern Korean bakeries''}---a claim loosely entailed by the cultural norm that Korean food is often adapted to local tastes (e.g., garlic bread with sugar), and thus accepted by norm-based verification. Yet the response strikes Korean speakers as off: the specific ingredient and pairing are not part of any recognizable Korean dessert practice---an instance of \textit{Cultural Incoherence} that norm-level entailment cannot adjudicate.

\paragraph{Claim verification fails on both surface and connotative grounds.}
We next evaluate the cultural claim extraction-and-verification pipeline. Across the 33,621 sentences in \dataset{}, the extractor identified cultural claims in 9,957 sentences, yielding 15,134 claims in total. Of these, 11,052 were not supported by Wikipedia evidence and 2,958 were not supported by Google Search evidence---reflecting Wikipedia's narrower coverage of the target cultures relative to the open web. Treating any sentence containing an unsupported claim as an error span and comparing against \dataset{}, both variants perform substantially worse than every LLM-judge in Table~\ref{tab:scores_span_detection}: the Wikipedia-based pipeline achieves an F1 of 0.20 (precision 0.17, recall 0.24), and the Google Search variant only 0.12 (precision 0.20, recall 0.09). The two variants trade off in opposite directions---Wikipedia over-flags claims for which it simply has no entry, while Google Search under-flags claims that are technically findable on the web but culturally misused---and neither approaches the performance of even the weakest direct LLM-judge.

More tellingly, 2,122 sentences contain a claim that the Search-based verifier marks as supported, yet \dataset{}'s annotators flag the surrounding span as culturally erroneous. A representative case: the claim \textit{``Han is a distinct Korean cultural concept referring to a collective feeling of grief and sorrow''} is correctly verified as true, but the response embeds it within a description of \textit{Korean funeral practices}, framing \textit{han} as a concept specifically about mourning the dead. While the literal definition is supported by external evidence, the connotative scope of \textit{han}---a diffuse, historical sentiment far broader than funerary grief---is misrepresented. Verification against atomic factual support has no mechanism to detect this kind of mismatch, which is precisely the territory where thick cultural errors live.

\begin{figure}[ht]
    \centering
    \includegraphics[width=0.5\linewidth]{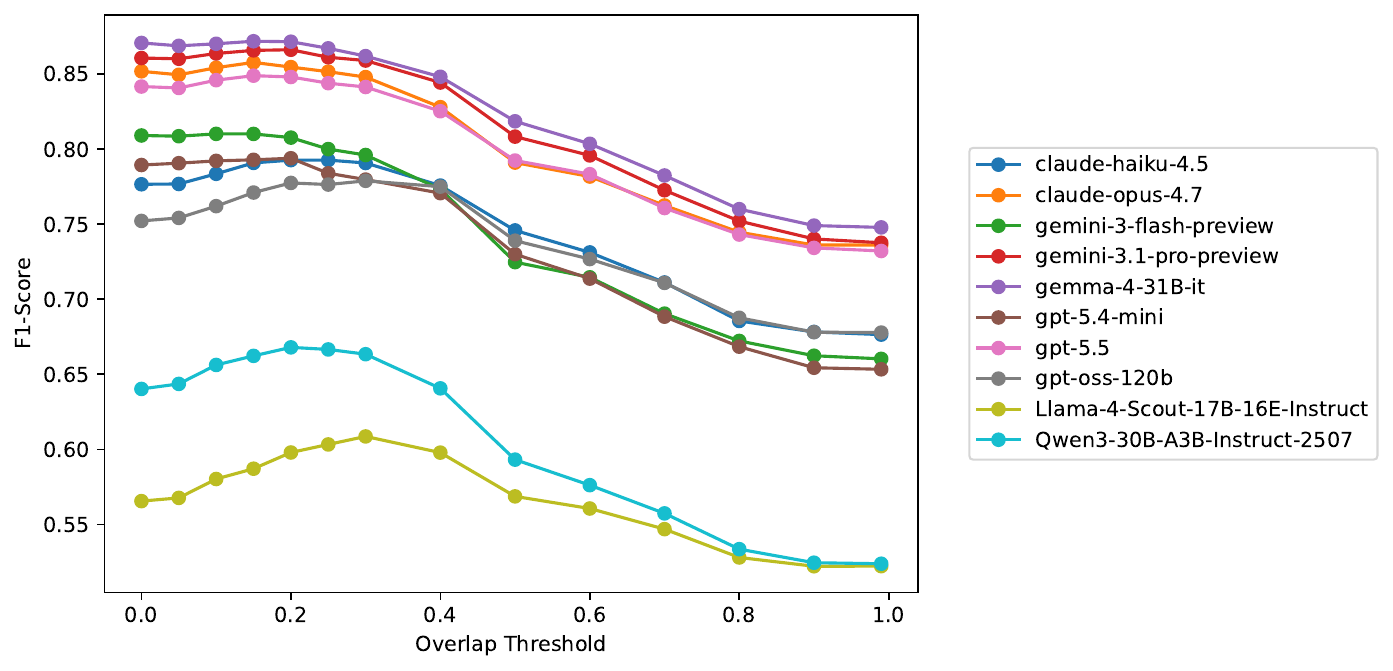}
    \caption{F1-scores of overlap-based evaluation against semantic-based evaluation with various IoU thresholds.}
    \label{fig:f1_threshold}
\end{figure}

\begin{figure}
    \centering
    \includegraphics[width=\textwidth]{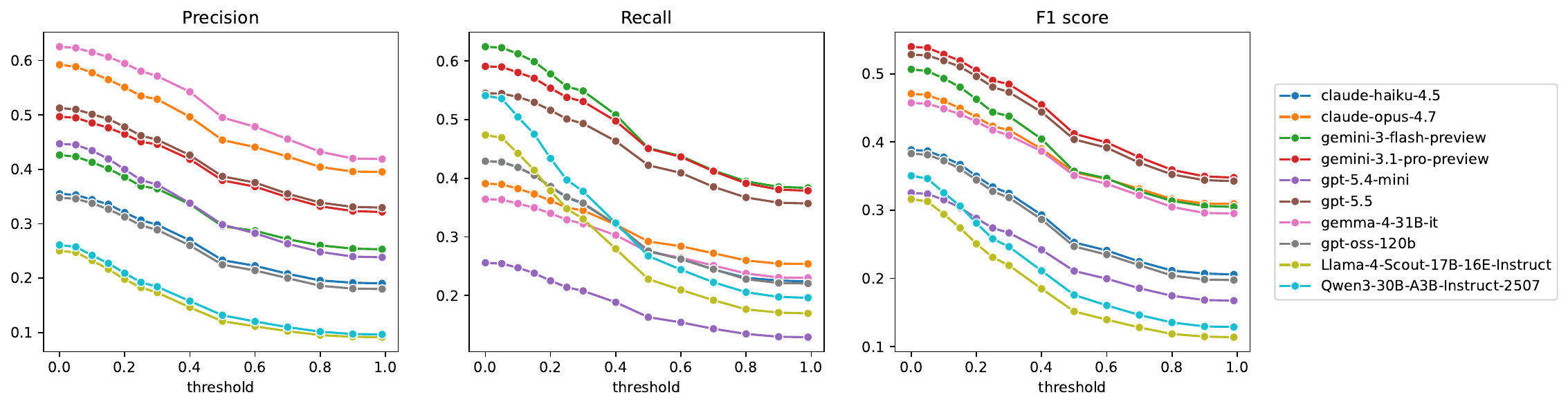}
    \caption{Span detection performance by IoU threshold.}
    \label{fig:span_detection_threshold}
\end{figure}

\begin{figure}
    \centering
    \includegraphics[width=\textwidth]{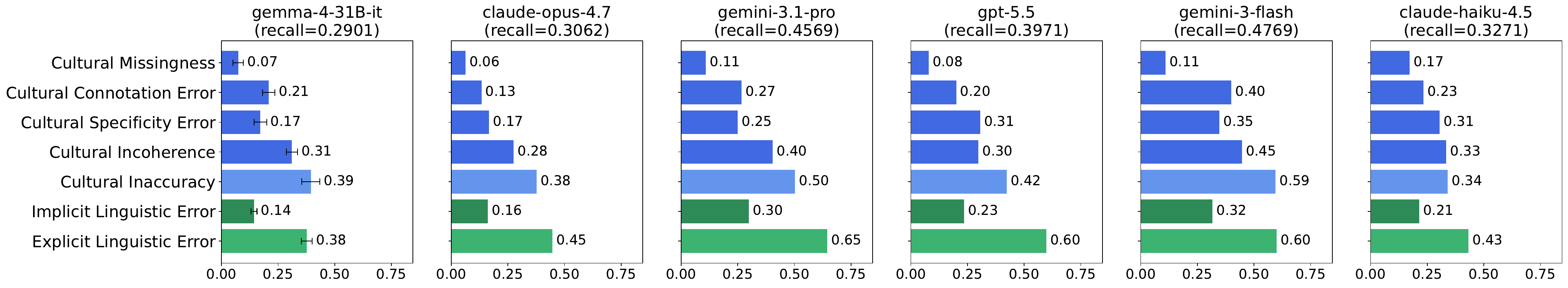}
    \caption{Category-wise recall of LLM-judges with top-6 F1 scores on the span detection of human-detected errors.}
    \label{fig:category_recall_span_detection_human}
\end{figure}

\subsection{Erroneous span detection}
\label{appendix:eval_details_span_detect}

\subsubsection{Evaluation setup}
\label{appendix:eval_details_span_detect_setup}
To evaluate the predicted spans, we use a word-level Intersection over Union (IoU) metric. A model's predicted span is considered a true positive if there exists a ground-truth span in our dataset such that the IoU between the predicted and true span exceeds a predefined strictness threshold $t$.

To empirically determine a well-calibrated threshold $t$ that accurately reflects semantic equivalence, we implement a two-step alignment process:
\begin{enumerate}
    \item Heuristic Classification: We compute a word-level IoU-based ``same/different'' classification, where an error prediction is deemed the ``same'' as the ground truth if the IoU $> t$.
    \item Semantic Oracle Classification: We employ gemma-4-31B-it \cite{google2026gemma4} as an independent semantic judge. Given a pair of error spans and their accompanying rationales (one from the dataset, one from the model prediction), the oracle determines whether both spans fundamentally point to the identical underlying cultural or linguistic issue.
\end{enumerate}
As shown in \autoref{fig:f1_threshold}, we sweep across values of $t$ to find the threshold that maximizes the F1-score between the heuristic classification and the semantic oracle classification (i.e., finding the IoU threshold that best approximates the LLM-judged semantic match). We repeat this calibration across all evaluated models and take the median threshold, which yields $t = 0.2$. Using this threshold ($t=0.2$), the IoU-based heuristic achieves strong alignment with the semantic oracle, yielding an average F1-score of 0.79 and a median F1-score of 0.80 across all models.

\subsubsection{Evaluation results}
\label{appendix:eval_details_span_detect_result}

\autoref{fig:span_detection_threshold} presents the span detection performance of LLM-judges across different IoU thresholds.
As thresholds become stricter, performance decreases; however, most models share a similar degradation trajectory, thereby preserving their relative order.

\autoref{fig:category_recall_span_detection_human} shows the overall and category-wise recall of LLM-judges in the span detection task on human-detected errors.

\input{sources/tab_scores_span_detection_small_models}

We additionally evaluate smaller models: \texttt{gemma-4-E2B-it} \citep{google2026gemma4}, \texttt{gemma-4-E4B-it} \citep{google2026gemma4}, and \texttt{Qwen3-4B-Instruct-2507} \citep{yang2025qwen3}. 
\autoref{tab:scores_span_detection_small_models} shows that these smaller models achieve lower scores than their larger counterparts.

\subsubsection{Results by Country and Language}

Table~\ref{tab:country_language_f1} reports span-level F1 by country and response language. These scores should be interpreted with caution, as the distribution and difficulty of annotated errors vary substantially across subsets. Consequently, differences in F1 may reflect both judge capability and subset composition, rather than differences in cultural understanding alone.

Averaged across the 14 evaluated models, F1 ranges from 0.214 for US-English to 0.419 for Bangladesh-Bengali. This variation may partly reflect differences in error composition: US-English has the highest proportion of thick errors, whereas Bangladesh-Bengali contains a larger proportion of explicit linguistic errors, which may be easier to detect. Across settings, the strongest performance is consistently achieved by frontier proprietary models (\texttt{gemini-3.1-pro-preview} and \texttt{gpt-5.5}). Meanwhile, region-specific models (\texttt{EXAONE-4.5-33B}, \texttt{A.X-4.0-Light}, \texttt{Qwen-SEA-LION-v4-32B-IT}, and \texttt{gemma2-9b-cpt-sahabatai-v1-instruct}) do not outperform the strongest general-purpose models, even when evaluating responses in their target languages.

We also compare recall on thick and thin errors.
Averaged across the 14 models, thick-error recall is lower than thin-error recall in all seven settings, with gaps from 0.031 to 0.097.
In Korea and Indonesia, recall is higher for English responses than for native-language responses for both error types, and the gap is larger for thick errors.
In Indonesia, for example, the English-native gap is 0.071 for thick errors and 0.032 for thin errors.
In Bangladesh, thick-error recall shows a similar English advantage (0.049), while thin-error recall is almost the same in the two languages.
This suggests that LLM judges detect thick errors more reliably in English, although differences in item difficulty across languages remain a confounding factor.

\begin{table*}[t]
\centering
\caption{Span-level F1 by country and response language on JuICE. The best score in each column is shown in bold.}
\label{tab:country_language_f1}
\vspace{8pt}
\resizebox{\textwidth}{!}{%
\begin{tabular}{l c cc cc cc}
\toprule
 & \textbf{US} & \multicolumn{2}{c}{\textbf{Korea}} & \multicolumn{2}{c}{\textbf{Indonesia}} & \multicolumn{2}{c}{\textbf{Bangladesh}} \\
\cmidrule(lr){2-2} \cmidrule(lr){3-4} \cmidrule(lr){5-6} \cmidrule(lr){7-8}
Model & English & Native & English & Native & English & Native & English \\
\midrule
gemini-3.1-pro-preview & 0.302 & \textbf{0.481} & \textbf{0.503} & 0.452 & \textbf{0.481} & \textbf{0.619} & \textbf{0.515} \\
gpt-5.5 & \textbf{0.337} & 0.478 & 0.478 & \textbf{0.465} & 0.422 & 0.614 & 0.498 \\
\midrule
claude-haiku-4.5 & 0.233 & 0.322 & 0.339 & 0.296 & 0.326 & 0.469 & 0.346 \\
claude-opus-4.7 & 0.215 & 0.405 & 0.426 & 0.336 & 0.370 & 0.583 & 0.461 \\
gemini-3-flash-preview & 0.277 & 0.427 & 0.474 & 0.425 & 0.454 & 0.580 & 0.454 \\
gpt-5.4-mini & 0.167 & 0.204 & 0.285 & 0.220 & 0.273 & 0.417 & 0.295 \\
\midrule
gemma-4-31B-it & 0.270 & 0.396 & 0.444 & 0.339 & 0.421 & 0.536 & 0.461 \\
gpt-oss-120b & 0.253 & 0.306 & 0.377 & 0.310 & 0.337 & 0.409 & 0.354 \\
Llama-4-Scout-17B-16E-Instruct & 0.165 & 0.238 & 0.259 & 0.220 & 0.262 & 0.341 & 0.246 \\
Qwen3-30B-A3B-Instruct-2507 & 0.184 & 0.268 & 0.303 & 0.260 & 0.278 & 0.362 & 0.283 \\
\midrule
EXAONE-4.5-33B & 0.147 & 0.235 & 0.237 & 0.178 & 0.229 & 0.221 & 0.250 \\
A.X-4.0-Light & 0.129 & 0.175 & 0.218 & 0.186 & 0.192 & 0.188 & 0.179 \\
Qwen-SEA-LION-v4-32B-IT & 0.182 & 0.272 & 0.300 & 0.263 & 0.287 & 0.359 & 0.276 \\
gemma2-9b-cpt-sahabatai-v1-instruct & 0.131 & 0.133 & 0.168 & 0.153 & 0.169 & 0.163 & 0.160 \\
\bottomrule
\end{tabular}%
}
\end{table*}

\subsection{Erroneous sentence classification}
\subsubsection{Sentence-level dataset construction}
To construct the sentence-level version of our dataset, we first split each response by newline characters (\texttt{\textbackslash n}). We then segment each line into sentences using \texttt{nltk.tokenize.sent\_tokenize}\footnote{\url{https://pypi.org/project/nltk/}} for English, Korean, and Indonesian, and \texttt{bnlp.NLTKTokenizer.sentence\_tokenize}\footnote{\url{https://pypi.org/project/bnlp-toolkit/}} for Bengali.

\subsubsection{Evaluation results}
\label{appendix: eval_result_classification}

\begin{figure}
    \centering
    \includegraphics[width=\textwidth]{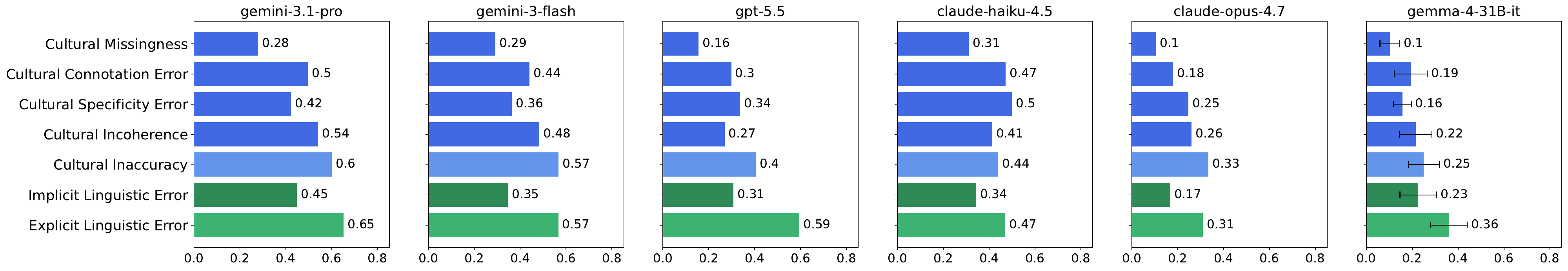}
    \caption{Category-wise recall of LLM-judges with top-6 F1 scores on the sentence classification task.}
    \label{fig:category_recall_classification}
\end{figure}

\autoref{tab:scores_classification} and \autoref{fig:category_recall_classification} show the performance and category-wise recall of LLM-judges in the sentence classification task.

\input{sources/tab_scores_classification}

\subsection{Guiding LLM-judges with thick taxonomy}
\input{sources/minipage_improvement_full}

\input{sources/tab_improvement_country_language}

\subsubsection{Setting}
\label{appendix:improvement_setting}
The prompts used for guiding LLM-judges with thick taxonomy are in Appendix \ref{appendix:improvement_tax}, \ref{appendix:improvement_fewshots}, and \ref{appendix:improvement_tax_shots}.
To strictly separate the few-shot demonstrations from the test set, we first sample one query from each of the three tasks. We then extract demonstrations from queries of the same type (i.e., variants involving different countries or languages), ensuring they are balanced across error categories. Finally, we exclude those specific query types from the test set to evaluate performance.

\subsubsection{Guiding \texttt{gemma-4-31B-it}}
\label{appendix:improvement_gemma}
\autoref{tab:improvement_full} reports the mean and standard deviation of span detection performance scores for \texttt{gemma-4-31B-it} across four trials with different instructions and demonstration sets.

\autoref{tab:improvement_country_language} presents the recall scores of \texttt{gemma-4-31B-it} broken down by country-language pair for the inference-time guidance experiment. Within each country-language pair, the scores consistently maintain the following order in almost all cases: \emph{Span detection} < \emph{+ Taxonomy Definition} < \emph{+ Taxonomy Examples} < \emph{+ Taxonomy Definition + Taxonomy Examples} (with the only exception being the US-en case). This demonstrates that our guidance methods are broadly effective across countries and languages.

\subsubsection{Guiding \texttt{gemini-3.1-pro-preview} and \texttt{gemini-3-flash-preview}}
\label{appendix:improvement_gemini}

To demonstrate the effectiveness of guiding LLM-Judges with the thick taxonomy on other frontier models beyond the \texttt{gemma-4-31B-it} model, we conduct additional experiments using \texttt{gemini-3.1-pro-preview} and \texttt{gemini-3-flash-preview}. For these models, we run a single trial per setting using the prompts in Appendix \ref{appendix:improvement_tax}, \ref{appendix:improvement_fewshots}, and \ref{appendix:improvement_tax_shots}.

\autoref{tab:improvement_gemini} presents the average gains achieved by the three taxonomy-guidance settings (\emph{+ Taxonomy Definition}, \emph{+ Taxonomy Examples}, and \emph{+ Taxonomy Definition + Taxonomy Examples}), compared to the baseline without guidance for both models. Consistent with the results from \texttt{gemma-4-31B-it}, precision decreased while recall increased in both models; notably, regarding category-wise recall, the gains are larger for thick errors than for thin ones.
These results suggest that taxonomy guidance helps LLM-judges detect broader and more nuanced cultural errors, albeit at the cost of reduced precision.

\input{sources/tab_improvement_gemini}

\section{Model inference}
\label{appendix:computing_resource}

\paragraph{Proprietary models.}
We use APIs and inference infrastructure from OpenAI~\footnote{\url{https://openai.com/}}, Google AI Studio~\footnote{\url{https://aistudio.google.com/}}, and
OpenRouter~\footnote{\url{https://openrouter.ai/}}.

\begin{itemize}
    \item \textbf{OpenAI.} We use the following model through OpenAI Platform API: \texttt{gpt-5.1-2025-11-13} \citep{openai2025gpt51}, \texttt{gpt-5-mini-2025-08-07} \citep{singh2026openaigpt5}, and \texttt{gpt-5.4-mini-2026-03-17} \citep{openai2026gpt54mini}.
    \item \textbf{{Google AI Studio.}} We use the following models through Google AI Studio API: \texttt{gemini-3-flash-preview} \citep{google2025gemini3flash} and \texttt{gemini-3.1-pro-preview} \citep{google2026gemini31pro}.
    \item \textbf{OpenRouter.} We use the following models through OpenRouter API: \texttt{claude-haiku-4.5} \citep{anthropic2025claudehaiku45}, \texttt{claude-opus-4.7} \citep{anthropic2026claudeopus47}, \texttt{gemini-2.5-flash} \citep{comanici2025gemini25}, \texttt{gemini-3-pro-preview} \citep{google2025gemini3pro}, \texttt{gpt-5.4-mini} \citep{openai2026gpt54mini}, and \texttt{gpt-5.5} \citep{openai2026gpt55}.
\end{itemize}

For the error annotation stage, we use \texttt{gemini-3-pro-preview} with a reasoning effort of \textit{low} and \texttt{gpt-5.1-2025-11-13} with a reasoning effort of \textit{medium}.

\paragraph{Open models.}
We use servers with 8 $\times$ 141GB H200 GPUs to run \texttt{Qwen3-235B-A22B-Instruct-2507}, and
servers with 4 $\times$ 80GB H100 GPUs to run
\texttt{gemma-4-31B-it} \citep{google2026gemma4},
\texttt{gpt-oss-120b} \citep{openai2025gptoss},
\texttt{Llama-4-Scout-17B-16E-Instruct} \citep{meta2025llama4scout}, and 
\texttt{Qwen3-30B-A3B-Instruct-2507} \citep{yang2025qwen3}.
We run the inference using vLLM \citep{kwon2023efficient} with the default parameters.

%% file: sources/tab_taxonomy_detail.tex
{\small
\begin{longtable}{
    @{}p{0.16\textwidth}
    p{0.05\textwidth}
    p{0.62\textwidth}@{}
}

\caption{Taxonomy for thick evaluations (full version).} \label{tab:taxonomy_detail} \\

\toprule
Category &  & \\ 
\midrule
\endfirsthead

\toprule
Category &  &  \\
\midrule
\endhead

\multirow[t]{2}{0.16\textwidth}{Explicit Linguistic Error} &
  Def. &
  \textbf{Surface-level, mechanical language mistakes.} This covers objective violations of language rules, including incorrect grammar, spelling errors, typos, literal translations that misuse vocabulary, or outputting text in the wrong language. These flaws are independent of social context.\\ \cmidrule{2-3}
 &
  \textit{E.g.} &
  \textbf{Error Span}: \textit{``fresh soy milk (kacang kedelai)''}\newline \textbf{Explanation}: In Indonesian, the common and correct term for soy milk is ``susu kedelai,'' not ``kacang kedelai.'' ``Kacang kedelai'' refers to the soybean itself (the bean), not the milk made from it, so this is a linguistic misuse of the phrase.\\ \midrule
\multirow[t]{2}{0.16\textwidth}{Implicit Linguistic Error} &
  Def. &
  \textbf{Failures in tone, register, or sociolinguistic nuance.} The text may be grammatically correct but reads unnaturally (e.g., ``awkward phrasing''), or completely misses the unspoken communication rules of the target language. \\ \cmidrule{2-3}
 &
  \textit{E.g.} &
  \textbf{Error Span}: \textit{``Nailed It!''}\newline
  \textbf{Explanation}: The English idiom 'Nailed It!' is colloquial and might not be immediately understood or carry the intended encouraging weight for a Korean student audience, even if they study English. A culturally localized product would likely use Korean phrases like 'Hwaiting!' (Fighting/Cheer up) or 'Hapgyeok' (Pass/Success) specifically relevant to exam culture. \\ \midrule
\multirow[t]{2}{0.16\textwidth}{Cultural Inaccuracy} &
  Def. &
  \textbf{Objectively incorrect verifiable facts about a culture.} Classify any factual errors regarding a specific country's artifacts, geography, local ingredients, traditions, historical events, institutions, or demographic makeup here. \\ \cmidrule{2-3}
 &
  \textit{E.g.} &
  \textbf{Error Span}: \textit{``environmental science''}\newline
  \textbf{Explanation}: Universitas Gadjah Mada (UGM) does not offer an undergraduate bachelor's degree program specifically in Environmental Science. The closest equivalent major available at the university is Environmental Geography. \\ \midrule
\multirow[t]{2}{0.16\textwidth}{Cultural Incoherence} &
  Def. &
  \textbf{A contextual clash of cultural elements.} The cultural elements mentioned are real, but combining them in the given setting, situation, or alongside other elements is jarring, awkward, or highly unnatural. \\ \cmidrule{2-3}
 &
  \textit{E.g.} & 
  \textbf{Error Span}:\textit{``making traditional snacks (e.g., *klepon* or *lumpia*)''}\newline
  \textbf{Explanation}: Making traditional snacks like klepon or lumpia feels out of place in an educational garden setting. It does not align with the other listed activities, which are strictly focused on farming and gardening. \\ \midrule
\multirow[t]{2}{0.16\textwidth}{Cultural Specificity Error} &
  Def. &
  \textbf{Misrepresenting the boundaries of a cultural concept.} This occurs when the model over-generalizes (applying a specific trait to an entire population via explicit bias/stereotyping), under-generalizes (treating a broad cultural norm as overly narrow), or conflates distinct regional variations into a single monolith. \\ \cmidrule{2-3}
 &
  \textit{E.g.} & 
  \textbf{Error Span}: \textit{``kegiatan keagamaan/pengajian''}\newline 
  \textbf{Explanation}: The use of the slash in 'kegiatan keagamaan/pengajian' incorrectly implies that 'kegiatan keagamaan' (religious activities) and 'pengajian' (Islamic recitation/study) are synonymous. 'Kegiatan keagamaan' is a broader term, and since Indonesia officially recognizes six religions, equating it specifically with an Islamic practice inaccurately assumes that the students are Muslim. \\ \midrule
\multirow[t]{2}{0.16\textwidth}{Cultural Connotation Error} &
  Def. &
  \textbf{Mishandling the implicit symbolic weight or emotional resonance of a concept.} This applies when the response fails to grasp deep-seated symbolic, spiritual, or sentimental meanings (e.g., treating a sacred object as a trivial commodity), or carries unintended negative baggage through subtle framing and word choice. \\ \cmidrule{2-3}
 &
  \textit{E.g.} & 
  \textbf{Error Span}: \textit{``The scent of jasmine, typically a harbinger of peace''}\newline
  \textbf{Explanation}: In Indonesian culture, especially Javanese (implied by 'Kakek' and 'pendopo'), the strong scent of jasmine ('melati') is rarely associated simply with 'peace'. Instead, it is heavily associated with mysticism, spirits, death, and the supernatural. While jasmine flowers are used in funeral rituals (like washing the body and scattering flowers), the sudden smell of jasmine is culturally interpreted as a sign of a spirit's presence, often creating a sense of spookiness or reverence rather than peacefulness. \\ \midrule
\multirow[t]{2}{0.16\textwidth}{Cultural Missingness} &
  Def. &
  \textbf{The absence of critical, non-negotiable cultural context required for the response to be functionally or contextually complete.} This applies only when the omission of specific traditions, norms, or nuances results in a ``hollow'' or ``generic'' output that fails to address the unique cultural identity of the subject. \\ \cmidrule{2-3}
 &
  \textit{E.g.} & 
  \textbf{Error Span}: \textit{``umat Hindu, Kristen, Buddha, dan kepercayaan lokal.''}\newline
  \textbf{Explanation}: The LLM lists some of the minority religions in Indonesia but omits Confucianism (Konghucu) and Catholicism (Katolik). Since the Indonesian government officially acknowledges six religions (Islam, Protestantism/Kristen, Catholicism, Hinduism, Buddhism, and Confucianism), it is important to include all of them for accuracy. \\ \midrule
\multirow[t]{2}{0.16\textwidth}{Logical Error} &
  Def. &
  \textbf{Pure reasoning failures independent of language or culture.} This includes internal contradictions, invalid deductions, or scientific/mathematical flaws that have nothing to do with cultural context or linguistic mechanics. \\ \cmidrule{2-3}
 &
  \textit{E.g.} & 
  \textbf{Error Span}: \textit{``watch sticker''}\newline
  \textbf{Explanation}: During an exam, candidates usually bring a wristwatch to keep time, not a sticker of a watch, which would be functionally useless for time management. This is likely a hallucination or a linguistic error confusing a functional 'watch' (timepiece) with a decorative item. \\ \bottomrule
\\
\end{longtable}
}

%% file: sources/tab_annotator_stats.tex
\begin{table}[h]
\centering
\caption{Annotator demographics.}
\label{tab:annotator_demographics}
\begin{tabular}{lrrrr}
\toprule
 & US & KR & ID & BD \\
\midrule
\textbf{No. of Annotators} & 10 & 10 & 10 & 14 \\
\midrule
\textbf{Gender (\%)} & & & & \\
Female & 70.0 & 90.0 & 40.0 & 7.1 \\
Male & 30.0 & 10.0 & 60.0 & 92.9 \\
\midrule
\textbf{Age (\%)} & & & & \\
-24 & 40.0 & 80.0 & 60.0 & 71.4 \\
25--34 & 20.0 & 20.0 & 30.0 & 21.4 \\
35--44 & 10.0 & -- & 10.0 & 7.1 \\
45+ & 30.0 & -- & -- & -- \\
\bottomrule
\end{tabular}
\end{table}

%% file: sources/tab_annotation_stats.tex
\begin{table}[ht]
    \caption{Annotation statistics.}
    \label{tab:annotation_stats}
    \centering
    \resizebox{\linewidth}{!}{%
    \begin{tabular}{@{}lrrrrrrr@{}}
    \toprule
         & US & KR & KR & ID & ID & BD & BD \\
         & en & ko & en & id & en & bn & en \\
    \midrule
        (1st stage) avg. number of annotated errors \\
        -- human & 2.44 & 1.93 & 2.02 & 3.46 & 2.02 & 2.92 & 1.00 \\
        \quad -- high criticality & 0.31 & 0.48 & 0.75 & 0.73 & 0.51 & 1.21 & 0.24 \\
        \quad -- middle criticality & 0.94 & 0.96 & 0.92 & 1.57 & 1.01 & 1.21 & 0.56 \\
        \quad -- low criticality & 1.20 & 0.49 & 0.35 & 1.16 & 0.50 & 0.50 & 0.20 \\
        -- gemini-3-pro & 1.55 & 4.90 & 4.80 & 3.88 & 4.38 & 9.47 & 4.99 \\
        -- gpt-5.1 & 1.65 & 3.13 & 2.49 & 3.78 & 2.29 & 6.57 & 2.91 \\
    \midrule
        (2nd stage) avg. ratio of agreement \\
        -- detected by another human annotator & 0.6398 & 0.7945 & 0.8182 & 0.7190 & 0.7644 & 0.8721 & 0.7300 \\
        -- detected by gemini-3-pro & 0.8409 & 0.9097 & 0.9102 & 0.8801 & 0.8909 & 0.9085 & 0.8768 \\
        -- detected by gpt-5.1 & 0.8629 & 0.9227 & 0.9118 & 0.8794 & 0.9009 & 0.9281 & 0.8727 \\
    \bottomrule
    \end{tabular}
    }
\end{table}

%% file: sources/tab_scores_span_detection_small_models.tex
\begin{table}
    \caption{Span detection performance of small models.}
    \label{tab:scores_span_detection_small_models}
    \centering
    \begin{tabular}{llll}
    \toprule
    Model & Precision & Recall & F1 score \\
    \midrule
    gemma-4-E2B-it & $0.2740_{\pm 0.0143}$ & $0.0452_{\pm 0.0256}$ & $0.0752_{\pm 0.0369}$ \\
    gemma-4-E4B-it & $0.3014_{\pm 0.0419}$ & $0.2322_{\pm 0.0549}$ & $0.2536_{\pm 0.0247}$ \\
    gemma-4-31B-it & $0.5947_{\pm 0.0443}$ & $0.3401_{\pm 0.0318}$ & $0.4300_{\pm 0.0170}$ \\ \midrule
    Qwen3-4B-Instruct-2507 & $0.1940_{\pm 0.0084}$ & $0.3831_{\pm 0.0470}$ & $0.2566_{\pm 0.0139}$ \\
    Qwen3-30B-A3B-Instruct-2507 & $0.2090_{\pm 0.0112}$ & $0.4338_{\pm 0.0352}$ & $0.2810_{\pm 0.0058}$ \\
    \bottomrule
    \end{tabular}
\end{table}

%% file: sources/tab_scores_classification.tex
\begin{table}[b]
    \caption{Sentence classification performance on \dataset{}.}
    \label{tab:scores_classification}
    \resizebox{\textwidth}{!}{
    \centering
    \begin{tabular}{lllll}
    \toprule
    Model & Accuracy & Precision & Recall & F1 score \\
    \midrule
    gemini-3.1-pro-preview & $0.8519$ & $0.6108$ & $0.5285$ & $0.5667$ \\
    gpt-5.5 & $0.8546$ & $0.6787$ & $0.3916$ & $0.4967$ \\
    \midrule
    claude-haiku-4.5 & $0.7805$ & $0.4018$ & $0.4059$ & $0.4038$ \\
    claude-opus-4.7 & $0.8481$ & $0.7714$ & $0.2427$ & $0.3693$ \\
    gemini-3-flash-preview & $0.8444$ & $0.5969$ & $0.4629$ & $0.5214$ \\
    gpt-5.4-mini & $0.8249$ & $0.6061$ & $0.1257$ & $0.2082$ \\
    \midrule
    gemma-4-31B-it & $0.8434_{\pm 0.0006}$ & $0.7310_{\pm 0.0649}$ & $0.2488_{\pm 0.0670}$ & $0.3623_{\pm 0.0563}$ \\
    gpt-oss-120b & $0.7866_{\pm 0.0196}$ & $0.4054_{\pm 0.0447}$ & $0.3116_{\pm 0.0472}$ & $0.3470_{\pm 0.0233}$ \\
    Llama-4-Scout-17B-16E-Instruct & $0.7750_{\pm 0.0296}$ & $0.3275_{\pm 0.0889}$ & $0.1428_{\pm 0.0787}$ & $0.1732_{\pm 0.0762}$ \\
    Qwen3-30B-A3B-Instruct-2507 & $0.5253_{\pm 0.1952}$ & $0.2415_{\pm 0.0364}$ & $0.6433_{\pm 0.2117}$ & $0.3374_{\pm 0.0185}$ \\
    \bottomrule
    \end{tabular}
    }
\end{table}

%% file: sources/minipage_improvement_full.tex
\begin{table}
    \centering
    \caption{Span detection performance of \texttt{gemma-4-31B-it} under different prompting conditions.}
    \label{tab:improvement_full}
    \begin{tabular}{llll}
        \toprule
        Prompt & F1 score & Precision & Recall \\
        \midrule
        Span detection & $0.4307_{\pm 0.0183}$ & $0.5976_{\pm 0.0503}$ & $0.3398_{\pm 0.0346}$ \\ \midrule[0.3pt]
        + Only Thin Examples & $0.4259_{\pm 0.0204}$ & $0.5800_{\pm 0.0325}$ & $0.3386_{\pm 0.0329}$ \\ \midrule[0.3pt]
        + Taxonomy Examples & $0.4345_{\pm 0.0308}$ & $0.5210_{\pm 0.0420}$ & $0.3792_{\pm 0.0591}$ \\ \midrule[0.3pt]
        + Taxonomy Definition & $0.4378_{\pm 0.0127}$ & $0.5490_{\pm 0.0457}$ & $0.3671_{\pm 0.0310}$ \\ \midrule[0.3pt]
        + Taxonomy Definition & \multirow{2}{*}{$0.4407_{\pm 0.0224}$} & \multirow{2}{*}{$0.5208_{\pm 0.0276}$} & \multirow{2}{*}{$0.3851_{\pm 0.0439}$} \\
        + Taxonomy Examples \\
        \bottomrule
    \end{tabular}
\end{table}

%% file: sources/tab_improvement_country_language.tex
\begin{table}
    \centering
    \caption{Recall of \texttt{gemma-4-31B-it} on span detection, broken down by country-language pair.}
    \label{tab:improvement_country_language}
    \begin{tabular}{lccccccc}
        \toprule
        Prompt & US-en & KR-ko & KR-en & ID-id & ID-en & BD-bn & BD-en \\
        \midrule
        Span detection & $0.2038$ & $0.3103$ & $0.3543$ & $0.2463$ & $0.3252$ & $0.4409$ & $0.3979$ \\ \midrule[0.3pt]
        + Taxonomy Examples & $0.2026$ & $0.3364$ & $0.4129$ & $0.2872$ & $0.3922$ & $0.4616$ & $0.4695$ \\ \midrule[0.3pt]
        + Taxonomy Definition & $0.2277$ & $0.3300$ & $0.3926$ & $0.2723$ & $	0.3721$ & $0.4549$ & $0.4339$ \\ \midrule[0.3pt]
        + Taxonomy Definition & \multirow{2}{*}{$0.2086$} & \multirow{2}{*}{$0.3419$} & \multirow{2}{*}{$0.4190$} & \multirow{2}{*}{$0.2880$} & \multirow{2}{*}{$0.3997$} & \multirow{2}{*}{$0.4691$} & \multirow{2}{*}{$0.4787$} \\
        + Taxonomy Examples \\
        \bottomrule
    \end{tabular}
\end{table}

%% file: sources/tab_improvement_gemini.tex
\begin{table}
    \centering
    \caption{Performance gains in span detection with guidance. The `$\Delta$ Recall of thick cultural errors' row represents the average gain across its four constituent categories.}
    \label{tab:improvement_gemini}
    \begin{tabular}{lcc}
        \toprule
         & gemini-3.1-pro-preview & gemini-3-flash-preview \\
        \midrule
        $\Delta$ F1 score & $-0.0091$ & $-0.0157$ \\
        $\Delta$ Precision & $-0.0424$ & $-0.0392$ \\
        $\Delta$ Recall & $+0.0488$ & $+0.0524$ \\
        $\Delta$ Recall for thick cultural errors & $+0.1216$ & $+0.1036$ \\
        $\Delta$ Recall for thin cultural errors & $+0.0955$ & $+0.0542$ \\
        $\Delta$ Recall for thick linguistic errors & $+0.0353$ & $+0.0517$ \\
        $\Delta$ Recall for thin linguistic errors & $-0.0170$ & $+0.0161$ \\
        \bottomrule
    \end{tabular}
\end{table}